\pdfoutput=1
\documentclass{article}

 \usepackage[nonatbib,final]{neurips_2019}

\usepackage[utf8]{inputenc} 
\usepackage[T1]{fontenc}    
\usepackage{hyperref}       
\usepackage{url}            
\usepackage{booktabs}       
\usepackage{nicefrac}       
\usepackage{microtype}      
\usepackage{floatrow}
\usepackage{subcaption}
\usepackage{enumitem}
\usepackage{multirow}

\usepackage{floatrow}
\newfloatcommand{capbtabbox}{table}[][\FBwidth]

\usepackage{amssymb,amsmath}
\usepackage{mathtools}
\mathtoolsset{showonlyrefs}
\usepackage{mathbbol}

\usepackage{mdframed}
\definecolor{theoremcolor}{rgb}{1.0, 1.0, 1.0}
\mdfsetup{
    backgroundcolor=theoremcolor,
    linewidth=0.5pt,
}

\newmdtheoremenv{proposition}{Proposition}
\newmdtheoremenv{lemma}{Lemma}

\DeclareMathOperator*{\argmax}{argmax}
\DeclareMathOperator*{\argmin}{argmin}
\DeclareMathOperator*{\dom}{dom}

\DeclareMathOperator*{\conv}{conv}

\def\cB{\mathcal{B}}
\def\cC{\mathcal{C}}

\def\cL{\mathcal{L}}
\def\cM{\mathcal{M}}
\def\cO{\mathcal{O}}
\def\cP{\mathcal{P}}
\def\cS{\mathcal{S}}

\def\cX{\mathcal{X}}
\def\cY{\mathcal{Y}}

\def\RR{\mathbb{R}}
\def\EE{\mathbb{E}}

\def\ones{\mathbf{1}}
\def\yhat{\widehat{y}}
\def\muv{\mu_\varphi}

\def\Ssq{S_{\text{sq}}}

\def\MAP{\text{MAP}}
\def\LMO{\text{LMO}}
\def\marginal{\text{marginal}}

\title{Structured Prediction with Projection Oracles}

\author{%
  Mathieu Blondel\\
  NTT Communication Science Laboratories\\
  Kyoto, Japan \\
  \texttt{mathieu@mblondel.org} \\
}

\begin{document}

\maketitle

\begin{abstract}
We propose in this paper a general framework for deriving loss functions for
structured prediction.
In our framework, the user chooses a convex set including
the output space and provides an oracle for \textit{projecting} onto that set.
Given that oracle, our framework automatically generates a corresponding convex
and smooth loss
function.
As we show, adding a projection as output layer provably makes the loss
smaller.
We identify the marginal polytope, the output space's convex hull, as the
best convex set on which to project.  However, because the projection onto the
marginal polytope can sometimes be expensive to compute, we allow to use
any convex superset instead, with potentially cheaper-to-compute
projection. 
Since efficient projection algorithms are available for numerous
convex sets, this allows us to construct loss functions for a variety of tasks. 
On the theoretical side, 
when combined with \textit{calibrated decoding},
we prove that our loss functions can be used as a consistent surrogate for 
a (potentially non-convex) target loss function of interest.
We demonstrate our losses on label ranking, ordinal regression and
multilabel classification, confirming the improved accuracy enabled by
projections.
\end{abstract}

\section{Introduction}

The goal of supervised learning is to learn a mapping that links an input to an
output, using examples of such pairs. This task is noticeably more
difficult when the output objects have a structure, i.e., when they are not
mere vectors. This is the so-called structured prediction setting
\cite{bakir_2007} and has numerous applications in natural language processing,
computer vision and computational biology. 

We focus in this paper on the surrogate loss framework,
in which a convex loss is used as a proxy for a (potentially non-convex) target
loss of interest.
Existing convex losses for structured prediction 
come with different trade-offs. 
On one hand, the structured perceptron 
\cite{structured_perceptron} and hinge \cite{structured_hinge} losses
only require access to a maximum a-posteriori (MAP) oracle for finding the
highest-scoring structure, while the
conditional random field (CRF) \cite{lafferty_2001} loss requires access to a
marginal inference oracle, for evaluating the expectation under a Gibbs
distribution. Since marginal inference is generally considered harder than MAP
inference, for instance containing \#P-complete counting problems,
this makes the CRF loss less widely applicable.
On the other hand, unlike the structured perceptron and hinge
losses, the CRF loss is smooth, which is crucial for fast convergence, and comes
with a probabilistic model, which is important for dealing with uncertainty.
Unfortunately, when combined with MAP decoding, 
these losses are typically inconsistent, meaning that their optimal estimator
does not converge to the target loss function's optimal estimator.
Recently, several works
\cite{ciliberto_2016,korba_2018,nowak_2018,luise_2019}
showed good results and obtained consistency guarantees by combining a simple
squared loss with \textbf{calibrated decoding}. Since these approaches only
require a decoding oracle at test time and no oracle at train time, this
questions whether structural information is even beneficial during training. 

In this paper, we propose loss functions for
structured prediction using a different kind of oracle: \textbf{projections}.
Kullback-Leibler projections onto various polytopes have been used to derive
online algorithms
\cite{helmbold_2009,yasutake_2011,online_submodular,ailon_2016} but it is not
obvious how to extract a loss from these works.
In our framework, the user chooses a convex set containing
the output space and provides an oracle for projecting onto that set.
Given that oracle, we automatically generate an associated loss function.
As we show, incorporating a projection as output layer provably makes the loss
smaller.
We identify the marginal polytope, the output space's convex hull, as the
best convex set on which to project.  However, because the projection onto the
marginal polytope can sometimes be expensive to compute, we allow to use
instead any convex superset, with potentially cheaper-to-compute
projection. 
When using the marginal polytope as the convex set, our loss comes
with an implicit probabilistic model.
Our contributions are summarized as follows:
\begin{itemize}[topsep=0pt,itemsep=2pt,parsep=2pt,leftmargin=10pt]

\item Based upon Fenchel-Young losses \cite{fy_losses,fy_losses_journal}, we
    introduce projection-based losses in a broad setting. We give numerous
    examples of useful convex polytopes and their associated projections.

\item We study the \textbf{consistency} w.r.t. a target loss of interest when
    combined with calibrated decoding, extending a recent analysis
    \cite{nowak_2019} to the more general projection-based losses.
    We exhibit a \textbf{trade-off} between computational cost and statistical
    estimation.

\item We demonstrate our losses on label ranking, ordinal regression and
    multilabel classification, confirming the improved accuracy enabled by
    projections.

\end{itemize}

\paragraph{Notation.}

We denote the probability simplex by $\triangle^p \coloneqq
\{q \in \RR_+^p \colon \|q\|_1 = 1\}$,
the domain of  
$\Omega \colon \RR^p \rightarrow \RR\cup\{\infty\}$ 
by $\dom(\Omega) \coloneqq \{u \in \RR^p \colon
\Omega(u) < \infty\}$,
the Fenchel conjugate of $\Omega$ by 
$\Omega^*(\theta) \coloneqq \sup_{u \in \dom(\Omega)} \langle u,
\theta \rangle- \Omega(u)$.
We denote $[k] \coloneqq \{1, \dots, k\}$.

\vspace{-0.15cm}
\section{Background and related work}
\label{sec:background}

\vspace{-0.15cm}
\paragraph{Surrogate loss framework.}

The goal of structured prediction is to learn a mapping
$f \colon \cX \to \cY$, from an input $x \in \cX$
to an output $y \in \cY$, minimizing the expected target risk
\begin{equation}
\cL(f) \coloneqq \EE_{(X,Y) \sim \rho} ~ L(f(X), Y),
\end{equation}
where $\rho \in \triangle(\cX \times \cY)$ is a typically unknown distribution
and $L \colon \cY \times \cY \to \RR_+$ is a potentially non-convex target loss.
We focus in this paper on surrogate methods, which attack the problem in two
main phases. During the training phase, the labels $y \in \cY$ are first mapped
to $\varphi(y) \in \Theta$ using an \textbf{encoding} or embedding function
$\varphi \colon \cY \to \Theta$.  In this paper, we focus on $\Theta \coloneqq
\RR^p$, but some works consider general Hilbert spaces
\cite{ciliberto_2016,korba_2018,luise_2019}.  In most cases, $\varphi(y)$ will
be a \textbf{zero-one encoding} of the parts of $y$, i.e., $\varphi(y) \in
\{0,1\}^p$.
Given a surrogate loss $S \colon \Theta \times \Theta \to \RR_+$, a model $g
\colon \cX \to \Theta$ (e.g., a neural network or a linear model) is then
learned so as to minimize the surrogate risk
\begin{equation}
\cS(g) \coloneqq \EE_{(X,Y) \sim \rho} ~ S(g(X), \varphi(Y)).
\end{equation}
This allows to leverage the usual empirical risk minimization framework in the
space $\Theta$.
During the prediction phase, given an input $x \in \cX$,
a model prediction $\theta = g(x) \in \Theta$ is
``pulled back'' to a valid output $\yhat \in \cY$ using a \textbf{decoding}
function $d \colon \Theta \to \cY$. This is summarized in the following diagram:
\begin{equation}
x \in \cX
\xrightarrow[\text{model}]{g} 
\theta \in \Theta
\xrightarrow[\text{decoding}]{d} 
\yhat \in \cY.
\label{eq:decoding}
\end{equation}
Commonly used decoders include the pre-image oracle
\cite{weston_2003,cortes_2005,kadri_2013}
$\theta \mapsto \argmin_{y
\in \cY} S(\theta, \varphi(y))$ and the maximum a-posteriori inference
oracle \cite{structured_perceptron,structured_hinge,lafferty_2001},
which finds the highest-scoring structure:
\begin{equation}
\MAP(\theta) \coloneqq 
\argmax_{y \in \cY} \langle \theta, \varphi(y) \rangle.
\label{eq:MAP}
\end{equation}
In the remainder of this paper, for conciseness, we will use
use $S(\theta, y)$ as a shorthand for $S(\theta, \varphi(y))$ but it is useful
to bear in mind that surrogate losses are always really defined over vector spaces.

\vspace{-0.15cm}
\paragraph{Examples of surrogate losses.}

We now review classical examples of loss functions that fall within that
framework.  The structured perceptron \cite{structured_perceptron} loss is
defined by
\begin{equation}
S_{\text{SP}}(\theta, y) \coloneqq
\max_{y' \in \cY} ~ \langle \theta, \varphi(y') \rangle - 
\langle \theta, \varphi(y) \rangle.
\label{eq:sp_loss}
\end{equation}
Clearly, it requires a MAP inference oracle
at training time in order to compute subgradients w.r.t. $\theta$. 
The structured hinge loss used by structured support vector machines
\cite{structured_hinge} is a simple variant of
\eqref{eq:sp_loss} using an additional loss term.
Classically, it is assumed that this term satisfies an affine decomposition,
so that we only need a MAP oracle.
The conditional random fields (CRF) \cite{lafferty_2001} loss, 
on the other hand, requires a so-called
marginal inference oracle \cite{wainwright_2008}, for evaluating the expectation
under the Gibbs distribution $p(y; \theta) \propto e^{\langle \theta, \varphi(y)
\rangle}$.  The loss and the oracle are defined by
\begin{equation}
S_{\text{crf}}(\theta, y) \coloneqq
\log \sum_{y' \in \cY} e^{\langle \theta, \varphi(y') \rangle} - 
\langle \theta, \varphi(y) \rangle
\quad \text{and} \quad
\marginal(\theta) 
\coloneqq \EE_{Y \sim p}[\varphi(Y)] \propto
\sum_{y \in \cY} e^{\langle \theta, \varphi(y) \rangle} \varphi(y).
\label{eq:marginal_inference}
\end{equation}
When $\varphi(y)$ is a zero-one encoding of the parts of $y$ (i.e., a bit
vector), $\marginal(\theta)$ can be interpreted as some marginal distribution
over parts of the structures.
The CRF loss is smooth and comes with a probabilistic model, 
but its applicability is hampered by the fact that
marginal inference is generally harder than MAP inference.
This is for instance the case for permutation-prediction problems, where
exact marginal inference is intractable
\cite{valiant1979complexity,taskar-thesis,petterson_2009} but MAP inference can
be computed exactly.

\vspace{-0.15cm}
\paragraph{Consistency.}

When working with surrogate losses, an important question is whether the
surrogate and target risks are consistent, that is, whether an estimator
$g^\star$ minimizing $\cS(g)$ produces an estimator $d \circ
g^\star$ minimizing $\cL(f)$.
Although this question has been widely studied in the
multiclass setting \cite{zhang_2004,bartlett_2006,tewari_2007,mroueh_2012} and in other
specific settings \cite{duchi_2010,ravikumar_2011}, it is only recently that it
was studied in a fully general structured prediction setting. 
The structured perceptron, hinge and CRF losses are generally not
consistent when using MAP as decoder $d$ \cite{nowak_2019}. 
Inspired by kernel dependency estimation
\cite{weston_2003,cortes_2005,kadri_2013},
several works \cite{ciliberto_2016,korba_2018,luise_2019} showed good empirical
results and proved consistency by combining a
squared loss $\Ssq(\theta, y) \coloneqq \frac{1}{2} \|\varphi(y) -
\theta\|^2_2$ with calibrated decoding (no oracle is needed during
training). A drawback of this loss, however, is that it does not make use of
the output space $\cY$ during training, ignoring precious structural
information. More recently, the consistency of the CRF loss in combination with
calibrated decoding was analyzed in \cite{nowak_2019}.

\vspace{-0.15cm}
\section{Structured prediction with projection oracles}

\begin{figure*}
\includegraphics[height=3.8cm]{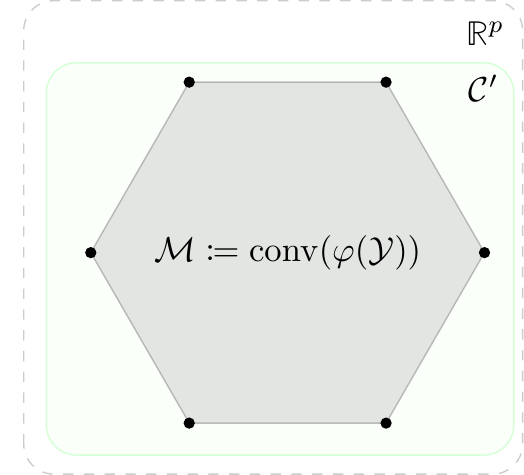}
\hspace{5pt}
\includegraphics[height=3.8cm]{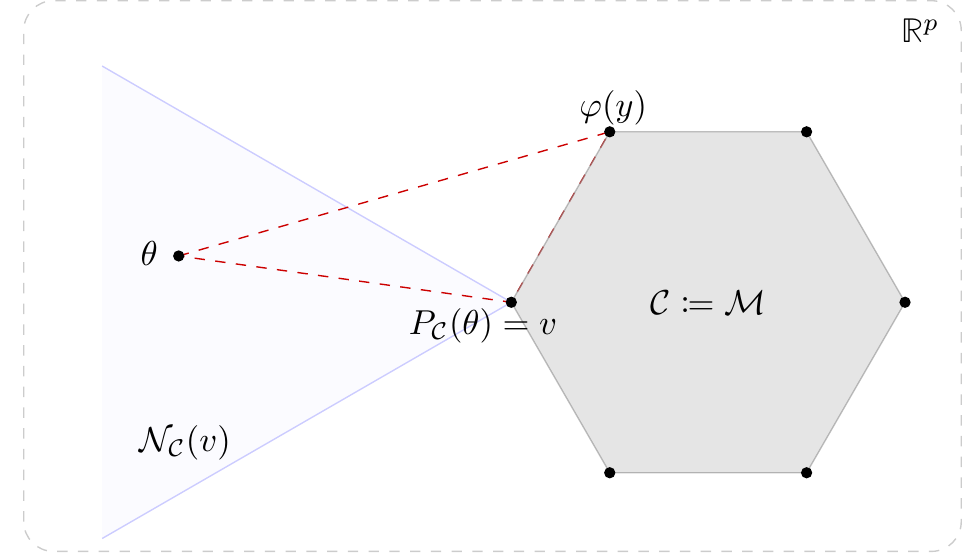}
\caption{Proposed framework in the Euclidean geometry.
\textbf{Left.}
Each black point represents the vector encoding $\varphi(y)$ of one possible
structure $y \in \cY$.
We require to choose a convex set $\cC$ including the encoded output
space, $\varphi(\cY)$. 
The best choice is $\cM$, the convex hull of $\varphi(\cY)$, but we can use any
superset $\cC'$ of it with potentially cheaper-to-compute projection.  Setting
$\cC = \RR^p$, our loss $S_\cC(\theta, y)$ (omitting the superscript $\Psi$)
recovers the squared loss (i.e., no projection).
\textbf{Right.}
When $\theta$ belongs to the interior of $\mathcal{N}_\cC(v)$, the normal
cone of $\cC$ at a vertex $v$, the projection $P_\cC(\theta) \coloneqq
\argmin_{u \in \cC} \|u - \theta\|_2$ hits the vertex $v$ and the angle formed
by $\theta$, $P_\cC(\theta)$ and $\varphi(y)$ is obtuse. In this case, 
$S_\cC(\theta, y)$ is a strict upper-bound for 
$\ell_\cC(\theta, y) \coloneqq \frac{1}{2} \|\varphi(y) - P_\cC(\theta)
\|^2_2$. When $\theta$ is not in the normal cone of $\cC$ at any vertex, then
the angle is right and the two losses coincide, $S_\cC(\theta, y) =
\ell_\cC(\theta, y)$.
}
\label{fig:projection}
\end{figure*}

\vspace{-0.15cm}
In this section, we build upon Fenchel-Young losses
\cite{fy_losses,fy_losses_journal} to derive a class of smooth loss functions
leveraging structural information through a different kind of oracle:
\textbf{projections}. Our losses are applicable to a large variety of tasks
(including permutation problems, for which CRF losses are intractable) and have
consistency guarantees when combined with calibrated decoding (cf. 
\S\ref{sec:consistency}).  

\vspace{-0.30cm}
\paragraph{Fenchel-Young losses.}

The aforementioned perceptron, hinge and CRF losses all belong to the
class of Fenchel-Young losses \cite{fy_losses,fy_losses_journal}.  The
Fenchel-Young loss generated by $\Omega$ is defined by 
\begin{equation}
S_\Omega(\theta, y) \coloneqq 
\Omega^*(\theta) + \Omega(\varphi(y)) - \langle \theta, \varphi(y) \rangle.
\label{eq:fy_loss}
\end{equation}
As shown in \cite{fy_losses,fy_losses_journal}, $S_\Omega(\theta, y)$
satisfies the following desirable properties:
\begin{itemize}[topsep=0pt,itemsep=2pt,parsep=2pt,leftmargin=10pt]

\item Non-negativity: $S_\Omega(\theta, y) \ge 0$,

\item Zero loss:
$S_\Omega(\theta, y) = 0 \Leftrightarrow 
\nabla \Omega^*(\theta) = \varphi(y)$,

\item Convexity: $S_\Omega(\theta, y)$ is convex in $\theta$,

\item Smoothness: If $\Omega$ is $\frac{1}{\beta}$-strongly
convex, then $S_\Omega(\theta, y)$ is $\beta$-smooth,

\item Gradient as residual (generalizing the squared loss): 
$\nabla_\theta S_\Omega(\theta, y) = \nabla \Omega^*(\theta) - \varphi(y)$.

\end{itemize}
In the Fenchel duality perspective, 
$\theta = g(x)$ belongs to the dual space $\dom(\Omega^*) = \Theta = \RR^p$ and
is thus unconstrained. This is convenient, as this places no restriction on the
model outputs $\theta = g(x)$. On the other hand, $\varphi(y)$ belongs to
the primal space $\dom(\Omega)$, which must include the encoded output space
$\varphi(\cY)$, i.e., $\varphi(\cY) \subseteq \dom(\Omega)$, and is typically
constrained. The gradient $\nabla \Omega^*$ is a mapping from $\dom(\Omega^*)$
to $\dom(\Omega)$ and $S_\Omega$ can be seen as loss with mixed arguments,
between these two spaces.  The theory of Fenchel-Young loss was recently
extended to infinite spaces in \cite{mensch_2019}.

\vspace{-0.30cm}
\paragraph{Projection-based losses.}

Let the Bregman divergence generated by $\Psi$ be defined as
$D_\Psi(u, v) \coloneqq \Psi(u) - \Psi(v) - \langle \nabla \Psi(v), u - v
\rangle$. The Bregman projection of $\nabla \Psi^*(\theta)$ onto a closed convex
set $\cC$ is
\begin{equation}
P_\cC^\Psi(\theta) \coloneqq 
\argmin_{u \in \cC} D_\Psi(u, \nabla \Psi^*(\theta)).
\label{eq:Bregman_projection}
\end{equation}
Intuitively, $\nabla \Psi^*$ maps the unconstrained predictions $\theta =
g(x)$ to $\dom(\Psi)$, ensuring that the Bregman projection is well-defined.
Let us define the Kullback-Leibler divergence by
$\text{KL}(u, v) \coloneqq 
\sum_i u_i \log \frac{u_i}{v_i} - \sum_i u_i + \sum_i v_i$. 
Two examples of generating function $\Psi$ are $\Psi(u) = \frac{1}{2} \|u\|_2^2$
with $\dom(\Psi) = \RR^p$ 
and $\nabla \Psi^*(\theta) = \theta$,
and $\Psi(u) = \langle u, \log u \rangle$ with $\dom(\Psi) = \RR_+^p$
and $\nabla \Psi^*(\theta) = e^{\theta - \mathbf{1}}$.
This leads to the Euclidean projection $\argmin_{u \in \cC}
\|u - \theta\|_2$ and the KL projection $\argmin_{u \in \cC} \text{KL}(u,
e^{\theta - \mathbf{1}})$, respectively.

Our key insight is to use a projection onto a chosen convex set $\cC$ as
\textbf{output layer}. If $\cC$ contains the encoded output
space, i.e., $\varphi(\cY) \subseteq \cC$, then $\varphi(y) \in \cC$ for any
ground truth $y \in \cY$. Therefore, if $\nabla \Psi^*(\theta) \not \in \cC$, then
$P_\cC^\Psi(\theta)$ is necessarily a better prediction than $\nabla
\Psi^*(\theta)$, since it is closer to $\varphi(y)$ in the sense of $D_\Psi$.
If $\nabla \Psi^*(\theta)$ already belongs to $\cC$, then 
$P_\cC^\Psi(\theta) = \nabla \Psi^*(\theta)$ and thus
$P_\cC^\Psi(\theta)$ is as good as $\nabla \Psi^*(\theta)$.
To summarize, we have $D_\Psi(\varphi(y), P_\cC^\Psi(\theta)) \le
D_\Psi(\varphi(y), \nabla \Psi^*(\theta))$ for all $\theta \in \Theta$ and $y
\in \cY$.  Therefore, it is natural to choose $\theta$ so as to minimize
the following \textbf{compositional} loss
\begin{equation}
    \ell_\cC^\Psi(\theta, y) \coloneqq D_\Psi(\varphi(y), P_\cC^\Psi(\theta)).
\end{equation}
Unfortunately, $\ell_\cC^\Psi$ is non-convex in $\theta$ in
general, and $\nabla_\theta \ell_\cC^\Psi(\theta, y)$ requires to compute the
Jacobian of $P_\cC^\Psi(\theta)$, which could be difficult, depending on $\cC$.
Other works have considered the output of an optimization
program as input to a loss \cite{stoyanov_2011,domke_2012,belanger_2017} but
these methods are non-convex too and typically require unrolling the program's
iterations. We address these issues, using Fenchel-Young losses.

\vspace{-0.30cm}
\paragraph{Convex upper-bound.}

We now set the generating function $\Omega$ of the Fenchel-Young loss
\eqref{eq:fy_loss} to
$\Omega = \Psi + I_\cC$, where $I_\cC$ denotes the indicator function of
$\cC$. We assume that $\Psi$ is Legendre
type \cite{rockafellar_1970,wainwright_2008}, meaning that it is strictly convex
and $\nabla \Psi$ explodes at the boundary of the interior of $\dom(\Psi)$.
This assumption is satisfied by both $\Psi(u) = \frac{1}{2} \|u\|_2^2$
and $\Psi(u) = \langle u, \log u \rangle$.
With that assumption, as shown in \cite{fy_losses,fy_losses_journal}, we obtain
$\nabla \Omega^*(\theta) = P_\cC^\Psi(\theta)$ for all $\theta \in \Theta$,
allowing us to use Fenchel-Young losses.  For
brevity, let us define the Fenchel-Young loss generated by 
$\Omega = \Psi + I_\cC$ as
\begin{equation}
S_\cC^\Psi(\theta, y) \coloneqq S_{\Psi + I_\cC}(\theta, y).
\label{eq:loss_shorthand}
\end{equation}
From the properties of Fenchel-Young losses, we have 
$S_\cC^\Psi(\theta, y) = 0 \Leftrightarrow P_\cC^\Psi(\theta) = \varphi(y)$
and $\nabla_\theta S_\cC^\Psi(\theta, y) = P_\cC^\Psi(\theta, y) - \varphi(y)$.
Moreover, as shown in \cite{fy_losses,fy_losses_journal}, $S_\cC^\Psi(\theta,
y)$ upper-bounds
$\ell_\cC^\Psi(\theta, y)$:
\begin{equation}
\ell_\cC^\Psi(\theta, y) \le S_\cC^\Psi(\theta, y)
\quad \forall \theta \in \Theta, y \in\cY.
\label{eq:upper_bound}
\end{equation}
Note that if $\cC = \dom(\Psi)$ (largest possible set),
then $S_\cC^\Psi(\theta, y) = D_\Psi(\varphi(y), \nabla \Psi^*(\theta))$.
In particular, with $\Psi = \frac{1}{2} \|\cdot\|_2^2$ and $\cC = \RR^p$,
$S_\cC^\Psi(\theta, y)$ recovers the squared loss
$\Ssq(\theta, y) = \frac{1}{2} \|\varphi(y) - \theta\|^2_2$.

\vspace{-0.30cm}
\paragraph{Choosing the projection set.}

Recall that $\cC$ should be a convex set such that $\varphi(\cY) \subseteq \cC$. 
The next new proposition, a simple consequence of \eqref{eq:fy_loss},
gives an argument in favor of using smaller sets.
\begin{proposition}{Using smaller sets results in smaller loss}
\label{prop:upper_bound}

Let $\cC,\cC'$ be two closed convex sets such that 
$\cC \subseteq \cC' \subseteq \dom(\Psi)$. Then,
\begin{equation}
S_\cC^\Psi(\theta, y) 
\le S_{\cC'}^\Psi(\theta, y) 
\quad \forall \theta \in \Theta, y \in \cY.
\end{equation}
\end{proposition}
As a corollary, combined with \eqref{eq:upper_bound}, we have 
\begin{equation}
\ell_\cC^\Psi(\theta, y) 
\le S_\cC^\Psi(\theta, y) 
\le D_\Psi(\varphi(y), \nabla \Psi^*(\theta))
\end{equation}
and in particular when $\Psi(u) = \frac{1}{2} \|u\|_2^2$,
noticing that $\Ssq = S_{\RR^p}^\Psi$, we have
\begin{equation}
\ell^\Psi_\cC(\theta, y) = \frac{1}{2} \|\varphi(y) - P_\cC^\Psi(\theta)\|^2_2
\le S_\cC^\Psi(\theta, y) 
\le \frac{1}{2} \|\varphi(y) - \theta\|^2_2
= \Ssq(\theta, y).
\end{equation}
Therefore, the Euclidean projection $P_\cC^\Psi(\theta)$ always achieves a
smaller squared loss than $\theta = g(x)$.
This is intuitive, as $\cC$ is a smaller region than
$\RR^p$ and $\cC$ is guaranteed to include the ground-truth $\varphi(y)$.
Our loss $S_\cC^\Psi$ is a convex and
\textbf{structurally informed} middle ground between $\ell_\cC^\Psi$ and $\Ssq$.

How to choose $\cC$?
The smallest convex set $\cC$ such that $\varphi(\cY) \subseteq \cC$ is the
\textbf{convex hull} of $\varphi(\cY)$
\begin{equation}
\cM \coloneqq \conv(\varphi(\cY)) \coloneqq 
\{ \EE_{Y \sim q}[\varphi(Y)] \colon q \in \triangle^{|\cY|} \} 
\subseteq \Theta.
\label{eq:marginal_polytope}
\end{equation}
When $\varphi(y)$ is a zero-one encoding of the parts of $y$, 
$\cM$ is also known as the \textbf{marginal polytope}
\cite{wainwright_2008},
since any point inside it can be interpreted as some marginal distribution over
parts of the structures.
The loss $S_\cC^\Psi$ with $\cC = \cM$ and $\Psi(u) = \frac{1}{2} \|u\|^2_2$ is
exactly the sparseMAP loss proposed in \cite{sparsemap}. 
More generally, we can use any \textbf{superset} $\cC'$ of $\cM$,
with potentially cheaper-to-compute projections.
For instance, when $\varphi(y)$ uses a zero-one encoding, the marginal polytope
is always contained in the \textbf{unit cube}, i.e., $\cM \subseteq [0,1]^p$,
whose projection is very cheap to compute.  We show in our experiments that even
just using the unit cube typically improves over the squared
loss. However, an advantage of using $\cC = \cM$ is that $P_\cM^\Psi(\theta)$
produces a convex combination of structures, i.e., an expectation.

\paragraph{Smoothness.}

The well-known equivalence between strong convexity of a function and the
smoothness of its Fenchel conjugate
implies that the following three statements are all equivalent:
\begin{itemize}[topsep=0pt,itemsep=2pt,parsep=2pt,leftmargin=10pt]

\item $\Psi$ is$\frac{1}{\beta}$-strongly convex w.r.t. a norm $\|\cdot\|$ over
    $\cC$,
\item $P_\cC^\Psi$ is $\beta$-Lipschitz continuous w.r.t. the dual norm
    $\|\cdot\|_*$ over $\RR^p$,
\item $S_\cC^\Psi$ is $\beta$-smooth in its first argument w.r.t. $\|\cdot\|_*$ over $\RR^p$.

\end{itemize}
With the Euclidean geometry, since $\Psi(u) = \frac{1}{2} \|u\|^2_2$ is
$1$-strongly-convex over $\RR^p$ w.r.t. $\|\cdot\|_2$, we have that $S_\cC^\Psi$
is $1$-smooth w.r.t. $\|\cdot\|_2$ \textbf{regardless} of $\cC$.
With the KL geometry, the situation is different.
The fact that $\Psi(u) = \langle u, \log u \rangle$ is $1$-strongly convex
w.r.t.  $\|\cdot\|_1$ over $\cC = \triangle^p$ is well-known (this is Pinsker's
inequality). The next proposition, proved in
\S\ref{appendix:proof_prop_strong_convexity},
shows that this straightforwardly extends to any bounded $\cC$ and that
the strong convexity constant is \textbf{inversely proportional} to the size of
$\cC$.
\begin{proposition}{Strong convexity of 
$\Psi(u) = \langle u, \log u \rangle$ over a bounded set}
\label{prop:strong_convexity}

Let $\cC \subseteq \RR_+^d$ and
$\beta \coloneqq \sup_{u \in \cC} \|u\|_1$.
Then, $\Psi$ is$\frac{1}{\beta}$-strongly convex w.r.t. $\|\cdot\|_1$ over $\cC$.
\end{proposition}
This implies that $S_\cC^\Psi$ is $\beta$-smooth w.r.t.\ $\|\cdot\|_\infty$.
Since smaller $\beta$ is smoother, this is another argument for
\textbf{preferring smaller sets} $\cC$.
With the best choice of $\cC = \cM$,
we obtain $\beta = \sup_{y \in \cY} \|\varphi(y)\|_1$. 

\paragraph{Computation.}

Assuming $\cC$ is compact (closed and bounded), the Euclidean projection can
always be computed using Frank-Wolfe or active-set algorithms, provided access
to a linear maximization oracle
$\LMO_\cC(v) \coloneqq \argmax_{u \in \cC} \langle u, v \rangle$.
Note that in the case $\cC = \cM$,
assuming that $\varphi$ is injective,
meaning that is has a left inverse,
MAP inference reduces to an LMO,
since $\MAP(\theta) = \varphi^{-1}(\LMO_\cM(\theta))$
(the LMO can be viewed as a linear program, whose solutions always hit a vertex
$\varphi(y)$ of $\cM$).
The KL projection is more problematic but Frank-Wolfe variants have been
proposed \cite{belanger_2013,krishnan_barrier}.
In the next section, we focus on examples of sets for which an efficient
\textbf{dedicated} projection oracle is available.

\vspace{-0.15cm}
\section{Examples of convex polytopes and corresponding projections}
\label{sec:polytope_examples}

\paragraph{Probability simplex.}

\begin{figure}[t]
\begin{subfigure}{.3\textwidth}
  \centering
  \includegraphics[height=3cm]{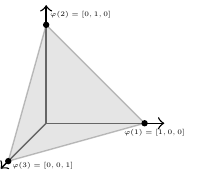}  
  \caption{Probability simplex}
\end{subfigure}
\begin{subfigure}{.3\textwidth}
  \centering
  \includegraphics[height=3cm]{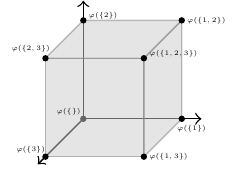}  
  \caption{Unit cube}
\end{subfigure}
\begin{subfigure}{.3\textwidth}
  \centering
  \includegraphics[height=3cm]{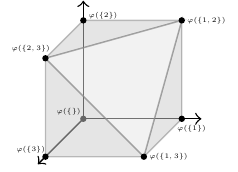}  
  \caption{Knapsack polytope}
  \label{fig:knapsack}
\end{subfigure}
\begin{subfigure}{.3\textwidth}
  \centering
  \includegraphics[height=3cm]{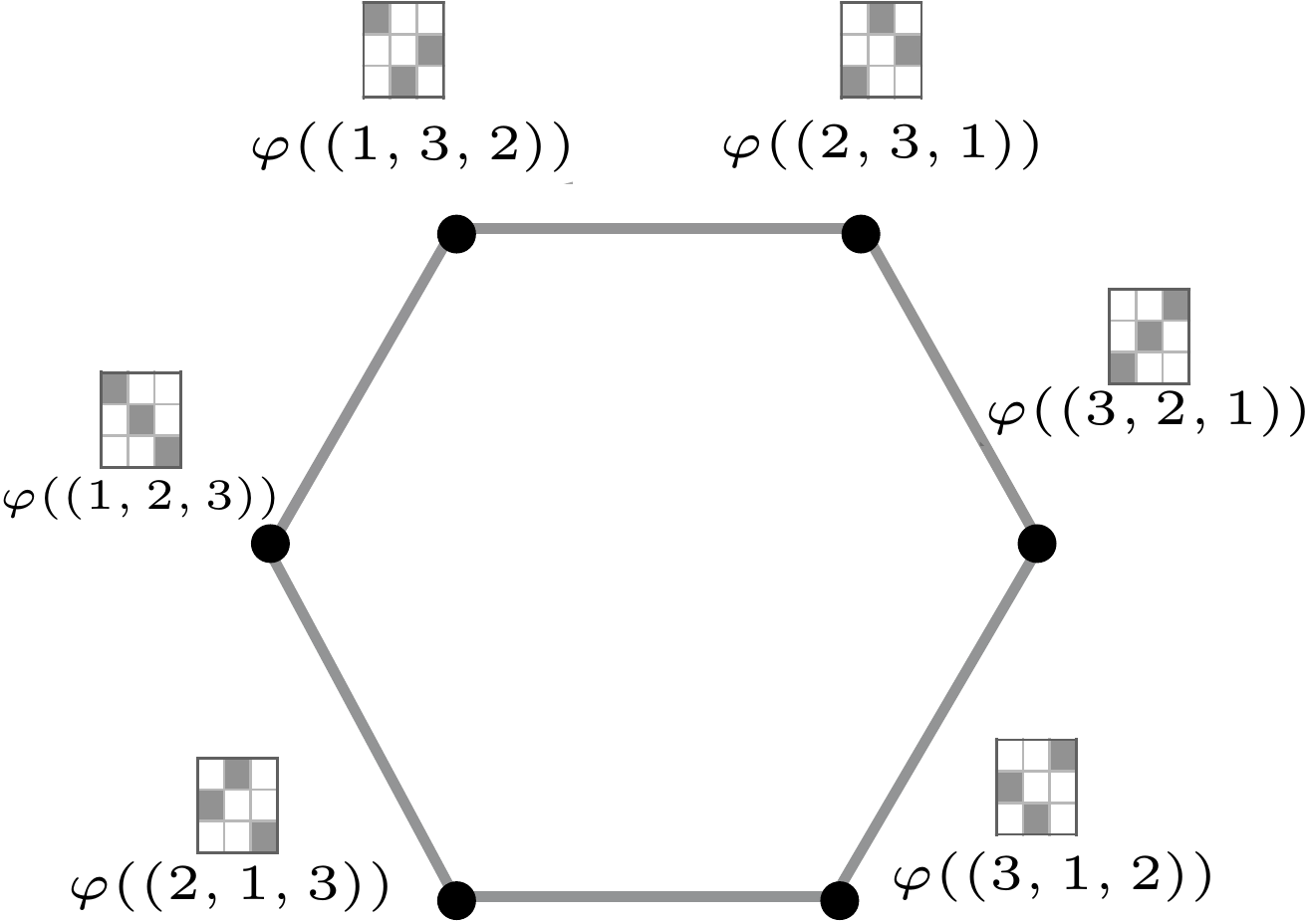}  
  \caption{Birkhoff polytope}
\end{subfigure}
\begin{subfigure}{.3\textwidth}
  \centering
  \includegraphics[height=3cm]{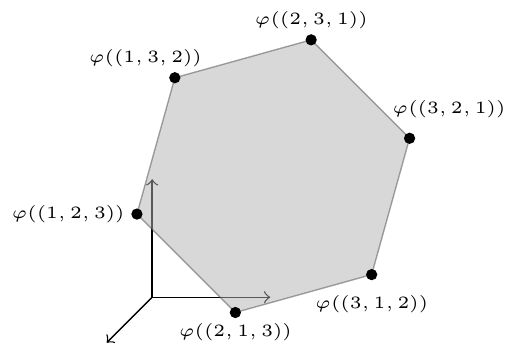}  
  \caption{Permutahedron}
\end{subfigure}
\begin{subfigure}{.3\textwidth}
  \centering
  \includegraphics[height=3cm]{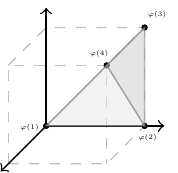}  
  \caption{Order simplex}
\end{subfigure}
\caption{Examples of convex polytopes.}
\label{fig:polytopes}
\end{figure}

For \textbf{multiclass classification}, we set  $\cY =
[k]$, where $k$ is the number of classes. With $\varphi(y) = \mathbf{e}_y$, 
the one-hot encoding of $y$, 
MAP inference \eqref{eq:MAP} becomes
$\MAP(\theta) = \argmax_{i \in [k]} \theta_k$. 
The marginal polytope defined in \eqref{eq:marginal_polytope} is now $\cM =
\triangle^k$, the probability simplex.  The Euclidean and KL
projections onto $\cC = \cM$ then correspond to
the sparsemax \cite{sparsemax} and softmax transformations. We therefore recover
the sparsemax and logistic losses as natural special cases of $S_\cC^\Psi$.
Note that, although the CRF loss \cite{lafferty_2001} also comprises the
logistic loss as a special case, it no longer coincides with our loss in the
structured case. 

\paragraph{Unit cube.}

For \textbf{multilabel classification}, we choose $\cY=2^{[k]}$, the powerset of
$[k]$.  Let us set $\varphi(y) = \sum_{i=1}^{|y|} \mathbf{e}_{y_i} \in
\{0,1\}^k$, the label indicator vector of $y$ (i.e., $\varphi(y)_i = 1$ if $i
\in y$ and $0$ otherwise).
MAP inference corresponds to predicting each label independently.
More precisely, for each label $i \in [k]$, 
if $\theta_i > 0$ we predict $i$, otherwise we do not.
The marginal polytope is now $\cM = [0,1]^k$, the unit cube.
Each vertex is in bijection with one possible subset of $[k]$.
The Euclidean projection of $\theta$ onto $\cM$ is equal to a
coordinate-wise clipping of $\theta$, i.e., 
$\max(\min(\theta_i, 1), 0)$ for all $i \in [k]$.
The KL projection is equal to $\min(1, e^{\theta_i - 1})$ for all $i \in [k]$.
More generally, whenever $\varphi$ for the task at hand uses a $0$-$1$ encoding,
we can use the unit cube as superset with computationally cheap projection. 

\paragraph{Knapsack polytope.}

We now set $\cY = \{y \in 2^{[k]} \colon l \le |y| \le u\}$, the subsets of
$[k]$ of bounded size. We assume $0 \le l \le u \le k$. This is useful for
\textbf{multilabel classification} with known lower bound $l \in \mathbb{N}$ and
upper bound $u \in \mathbb{N}$ on the number of labels per sample. 
Setting again $\varphi(y) =
\sum_{i=1}^{|y|} \mathbf{e}_{y_i} \in \{0,1\}^k$, MAP inference is equivalent to
the integer linear program $\argmax_{\varphi(y) \in \{0,1\}^k} \langle
\theta, \varphi(y) \rangle$ s.t. $l \le \langle \varphi(y), \ones \rangle \le
u$. Let $\pi$ be a permutation sorting $\theta$ in descending order. An optimal
solution is
\begin{equation}
\varphi(y)_i = \left\{
   \begin{array}{ll}
       1 & \text{if } l > 0 \text{ and } i \in \{\pi_1, \dots, \pi_l\}, \\
       1 & \text{else if } i \in \{\pi_1, \dots, \pi_u\} \text{ and } \theta_i
       > 0, \\
       0 & \text{else.}
   \end{array}\right.
\end{equation}
The marginal polytope is an instance of knapsack polytope \cite{almeida_2013}.
It is equal to
$\cM = \{ \mu \in [0,1]^k \colon l \le \langle \mu, \ones \rangle \le u \}$
and is illustrated in Figure \ref{fig:knapsack} with $k=3$, $l=0$ and $u=2$
(i.e., we keep all elements of $2^{[3]}$ except $\{1,2,3\}$).
The next proposition, proved in \S\ref{appendix:proj_budget},
shows how to efficiently project on $\cM$. 
\begin{proposition}{Efficient Euclidean and KL projections on $\cM$}
\label{prop:projection_budget}

\begin{itemize}[topsep=0pt,itemsep=2pt,parsep=2pt,leftmargin=10pt]
\item Let $\nu$ be the projection of $\nabla \Psi^*(\theta)$ onto the unit
    cube (cf.\ ``unit cube'' paragraph).
\item If $l \le \langle \nu, \ones \rangle \le u$, then $\nu$ is optimal.
\item Otherwise, return the projection of $\nabla \Psi^*(\theta)$ onto
$\{ \mu \in [0,1]^k \colon \langle \mu, \ones \rangle = m \}$,
where $m = u$ if $\langle \nu, \ones \rangle > u$ and $m=l$ otherwise.
\end{itemize}
\end{proposition}
The total cost is $O(k)$ in the Euclidean case and $O(k \log k)$ in the KL case
(cf. \S\ref{appendix:proj_budget} for details).



\paragraph{Birkhoff polytope.}

We view \textbf{ranking} as a structured prediction problem and let $\cY$ be the
set of permutations $\pi$ of $[k]$. Setting $\varphi(\pi) \in \{0,1\}^{k \times
k}$ as the permutation matrix associated with $\pi$, MAP inference becomes the
linear assignment problem $\MAP(\theta) = \argmax_{\pi \in \cY} \sum_{i=1}^k
\theta_{i, \pi_i}$ and can be computed exactly using the Hungarian algorithm
\cite{hungarian}.  The marginal polytope $\cM$ becomes the Birkhoff polytope
\cite{birkhoff}, the set of doubly stochastic matrices
\begin{equation}
\cM = \{P \in \RR^{k \times k} \colon P^\top \mathbf{1} = 1,
P \mathbf{1} = 1, 0 \le P \le 1 \}.
\end{equation}
Noticeably, marginal inference is known to be
\#P-complete \cite[\S3.5]{valiant1979complexity,taskar-thesis},
since it corresponds to computing a matrix permanent.
In contrast, the KL projection on the Birkhoff polytope can be computed
using the Sinkhorn algorithm \cite{sinkhorn,cuturi_2013}. The Euclidean
projection can be computed using Dykstra's algorithm \cite{rot_mover} or dual
approaches \cite{blondel_2018}. For both projections, the cost of
obtaining an $\epsilon$-precise solution is $O(k^2/\epsilon)$.
To obtain cheaper projections,
we can also use \cite{blondel_2018,nowak_2019} the set $\triangle^{k \times k}$
of row-stochastic matrices, a strict superset of the Birkhoff polytope and
strict subset of the unit cube
\begin{equation}
[0,1]^{k \times k} \supset
\triangle^{k \times k} \coloneqq
\triangle^k \times \dots \times \triangle^k =
\{P \in \RR^{k \times k} \colon P^\top \mathbf{1} = 1, 0 \le P \le 1 \}
\supset \cM.
\end{equation}
Projections onto $\triangle^{k \times k}$ reduce to $k$ row-wise projections
onto $\triangle^k$, for a \textit{worst-case} total cost of $O(k^2 \log k)$
in the Euclidean case and $O(k^2)$ in the KL case.

\paragraph{Permutahedron.}

We again consider \textbf{ranking} and let $\cY$ be the set of permutations
$\pi$ of $[k]$ but use a different encoding. This time, we
define $\varphi(\pi) = (w_{\pi_1}, \dots, w_{\pi_k}) \in \RR^k$, where $w \in
\RR^k$ is a prescribed vector of weights, which without loss of generality,
we assume sorted in descending order. 
MAP inference becomes $\MAP(\theta) = \argmax_{\pi \in \cY}
\sum_{i=1}^k \theta_i w_{\pi_i}$, whose solution is
the inverse of the permutation sorting $\theta$ in descending order, and can be
computed in $O(k \log k)$ time \cite{fy_losses_journal}.
When $w = (k, \dots, 1)$,
which we use in our experiments, $\cM$ is known
as the permutahedron and $\MAP(\theta)$ is equal to the ranks of $\theta$
(i.e., the inverse of $\text{argsort}(\theta)$). For arbitrary $w$, we follow
\cite{projection_permutahedron} and call $\cM$ the permutahedron induced by
$w$.  Its vertices correspond to the permutations of $w$.
Importantly, the Euclidean projection onto $\cM$ reduces to
sorting, which takes $O(k \log k)$, followed by isotonic regression, which takes
$O(k)$ \cite{zeng_2014,orbit_regul}. Bregman projections reduce to isotonic
\textit{optimization} \cite{projection_permutahedron}.

\paragraph{Order simplex.}

We again set $\cY = [k]$ but now consider the \textbf{ordinal regression}
setting, where there is an intrinsic order $1 \prec \dots \prec k$. We need to
use an encoding $\varphi$ that takes into account that order.  Inspired by the
all-threshold method \cite{pedregosa_2017,nowak_2019}, we set $\varphi(y) =
\sum_{1 \le i < y \le k} \mathbf{e}_i \in \RR^{k-1}$.
For instance, with $k=4$, we have $\varphi(1) = [0, 0, 0]$, $\varphi(2) = [1,
0, 0]$, $\varphi(3) = [1, 1, 0]$ and $\varphi(4) = [1,1,1]$. 
This encoding is also motivated by the fact that it enables consistency w.r.t.
the absolute loss (\S\ref{appendix:target_losses}).
As proved in \S\ref{appendix:vertices_order_simplex}, with that encoding, the
marginal polytope becomes the order simplex \cite{grotzinger_1984}.
\begin{proposition}{Vertices of the order simplex}
\label{prop:order_simplex}
\begin{equation}
\cM = \conv(0, e_1, e_1 + e_2, \dots, e_1 + \dots + e_{k-1})
= \{\mu \in \RR^{k-1} \colon 1 \ge \mu_1 \ge \mu_2 \ge \dots \ge \mu_{k-1} \ge 0 \}
\end{equation}
\end{proposition}
Note that without the upper bound on $\mu_1$, the resulting set is known as
monotone nonnegative cone \cite{boyd_2004}. The scores
$(\langle \theta, \varphi(y) \rangle)_{y=1}^k$ can be calculated using 
a cumulated sum in $O(k)$ time and therefore so do MAP and marginal inferences.
The Euclidean projection is equivalent to isotonic regression with lower and
upper bounds, which can be computed in $O(k)$ time \cite{best_1990}.

\vspace{-0.15cm}
\section{Consistency analysis of projection-based losses}
\label{sec:consistency}

We now study the consistency of $S_\cC^\Psi$ as a
proxy for a possibly non-convex target loss $L \colon \cY \times \cY \to
\RR_+$.

\paragraph{Affine decomposition.}

We assume that the target loss $L$ satisfies the
decomposition
\begin{equation}
L(\yhat, y) = \langle \varphi(\yhat), V \varphi(y) + b \rangle + c(y).
\label{eq:loss_decomp}
\end{equation}
This is a slight generalization of the decomposition of \cite{ciliberto_2016},
where we used an affine map $u \mapsto V u + b$ instead of a linear one and
where we added the term $c \colon \cY \to \RR$, which is independent of $\yhat$.
This modification allows us to express certain losses $L$ using a zero-one
encoding for $\varphi$ instead of a signed encoding \cite{nowak_2019}. The
latter is problematic when using KL projections and does not lead to
sparse solutions with Euclidean projections.
Examples of target losses satisfying \eqref{eq:loss_decomp} are discussed in
\S\ref{appendix:target_losses}.

\paragraph{Calibrated decoding.}

A drawback of the classical inference pipeline \eqref{eq:decoding} with decoder
$d = \MAP$ is that it is oblivious to the target loss $L$. In this paper, we
propose to use instead
\begin{equation}
x \in \cX
\xrightarrow[\text{model}]{g} 
\theta \in \Theta = \RR^p
\xrightarrow[\text{projection}]{P_\cC^\Psi} 
u \in \cC
\xrightarrow[\text{calibrated decoding}]{\yhat_L} 
\yhat \in \cY,
\label{eq:proposed_pipeline}
\end{equation}
where we define the decoding calibrated for the loss $L$ by
\begin{equation}
\yhat_L(u) 
\coloneqq \argmin_{y' \in \cY} \langle \varphi(y'), V u + b \rangle
= \MAP(-V u - b).
\label{eq:calibrated_decoding}
\end{equation}
Under the decomposition \eqref{eq:loss_decomp}, calibrated decoding therefore
reduces to MAP inference with pre-processed input. 
It is a ``rounding'' to $\cY$ of the projection $u =
P_\cC^\Psi(\theta) \in \cC$, that takes into account the loss $L$.  
Recently, \cite{ciliberto_2016,korba_2018,nowak_2018,luise_2019} used similar
calibrated decoding in conjunction with a squared loss (i.e., without an
intermediate layer) and \cite{nowak_2019} used it with a CRF loss (with marginal
inference as intermediate layer). To our knowledge, we are the first to use a
\textbf{projection layer} (in the Euclidean or KL senses) as an intermediate
step.

\paragraph{Calibrating target and surrogate excess risks.}

Given a (typically unknown) joint distribution $\rho \in \triangle(\cX \times
\cY)$, let us define the target risk of $f \colon \cX \to \cY$ 
and the surrogate risk of $g \colon \cX \to \Theta$ by
\begin{equation}
\cL(f) \coloneqq \EE_{(X,Y) \sim \rho} ~ L(f(X), Y)
\quad \text{and} \quad
\cS_\cC^\Psi(g) \coloneqq \EE_{(X,Y) \sim \rho} ~ S_\cC^\Psi(g(X), Y).
\end{equation}
The quality of estimators $f$ and $g$ is measured in terms of the \emph{excess}
of risks
\begin{equation}
\delta \cL(f) \coloneqq \cL(f) - \inf_{f'\colon \cX \to \cY} \cL(f')
\quad \text{and} \quad
\delta \cS_\cC^\Psi(g) \coloneqq \cS_\cC^\Psi(g) - \inf_{g' \colon \cX \to
\Theta} \cS_\cC^\Psi(g').
\end{equation}
The following proposition shows that $\delta \cL(f)$ and $\delta
\cS_\cC^\Psi(g)$ are calibrated when using our proposed inference pipeline
\eqref{eq:proposed_pipeline}, i.e., when $f = \yhat_L \circ P_\cC^\Psi \circ g$.
\begin{proposition}{Calibration of target and surrogate excess risks}
\label{prop:calibration}

Let $S_\cC^\Psi(\theta, y)$ and $L(\yhat, y)$ be defined as in
\eqref{eq:loss_shorthand} and \eqref{eq:loss_decomp}, respectively.  Assume
$\Psi$ is $\frac{1}{\beta}$-strongly convex w.r.t. $\|\cdot\|$ over $\cC$,
Legendre-type, and $\cC$ is a closed convex set such that $\varphi(\cY)
\subseteq \cC \subseteq \dom(\Psi)$.  Let $\sigma \coloneqq \sup_{\yhat \in \cY}
\|V^\top \varphi(\yhat)\|_*$, where $\|\cdot\|_*$ is the dual norm of $\|\cdot\|$.
Then,
\begin{equation}
\forall g \colon \cX \to \Theta: \quad
\frac{\delta \cL(\yhat_L \circ P_\cC^\Psi \circ g)^2}{8 \beta \sigma^2}
\le \delta \cS_\cC^\Psi(g).
\end{equation}
\end{proposition}
The proof, given in \S\ref{appendix:proof_prop_calibration}, is based on the
calibration function framework of
\cite{osokin_2017} and extends a recent analysis \cite{nowak_2019} to
projection-based losses.  Our proof covers Euclidean projection losses, not
covered by the previous analysis.
Proposition \ref{prop:calibration} implies Fisher consistency, i.e.,
$\cL(\yhat_L \circ P_\cC^\Psi \circ g^\star) = \cL(f^\star)$, where $f^\star
\coloneqq \argmin_{f \colon \cX \to \cY} \cL(f)$ and 
$g^\star \coloneqq \argmin_{g \colon \cX \to \Theta} \cS_\cC^\Psi(g)$.
Consequently, any optimization algorithm converging to $g^\star$ will also
recover an optimal estimator $\yhat_L \circ P_\cC^\Psi \circ g^\star$
of $\cL$. 
Combined with Propositions \ref{prop:upper_bound} and
\ref{prop:strong_convexity}, Proposition \ref{prop:calibration} suggests a
\textbf{trade-off} between computational cost and statistical estimation, larger
sets $\cC$ enjoying cheaper-to-compute projections but leading to slower rates.

\vspace{-0.15cm}
\section{Experimental results}

We present in this section our empirical findings on three tasks: label ranking,
ordinal regression and multilabel classification.  In all cases, we use a linear
model $\theta = g(x) \coloneqq W x$ and solve $\frac{1}{n} \sum_{i=1}^n
S_\cC^\Psi(W x_i, y_i) + \frac{\lambda}{2} \|W\|^2_F$ by L-BFGS, choosing
$\lambda$ against the validation set.  A Python implementation is available at 
\url{https://github.com/mblondel/projection-losses}.

\vspace{-0.15cm}
\paragraph{Label ranking.}

\begin{figure*}
\begin{floatrow}
\capbtabbox{%
\centering
\small
\begin{tabular}{cccccc}
\toprule
Projection & $\triangle^{k \times k}$ & $\RR^{k \times k}_+$ & $[0,1]^{k \times k}$ & $\triangle^{k \times k}$ & $\cM$ \\
Decoding & $\triangle^{k \times k}$ & $\cM$ & $\cM$ & $\cM$ & $\cM$ \\
\midrule
Authorship & 5.84 & \textbf{5.10} & 5.62 & 5.84 & \textbf{5.10} \\
Glass & 5.43 & 5.81 & 5.94 & 5.68 & \textbf{4.65} \\
Iris & 11.11 & 18.52 & 4.44 & \textbf{1.48} & 2.96 \\
Vehicle & 7.57 & 8.46 & 7.43 & 7.21 & \textbf{6.25} \\
Vowel & 9.50 & 9.40 & 9.42 & 9.28 & \textbf{9.17} \\
Wine & 4.32 & \textbf{1.85} & \textbf{1.85} & \textbf{1.85} & \textbf{1.85} \\
\bottomrule
\end{tabular}    
}{\caption{\label{table:label_ranking_KL}Same as Table
\ref{table:label_ranking_euc} but with KL projections instead.}}%
\ffigbox[\FBwidth][\FBheight][b]{%
\hspace{-.1cm}%
\centering%
\footnotesize \includegraphics[scale=0.30]{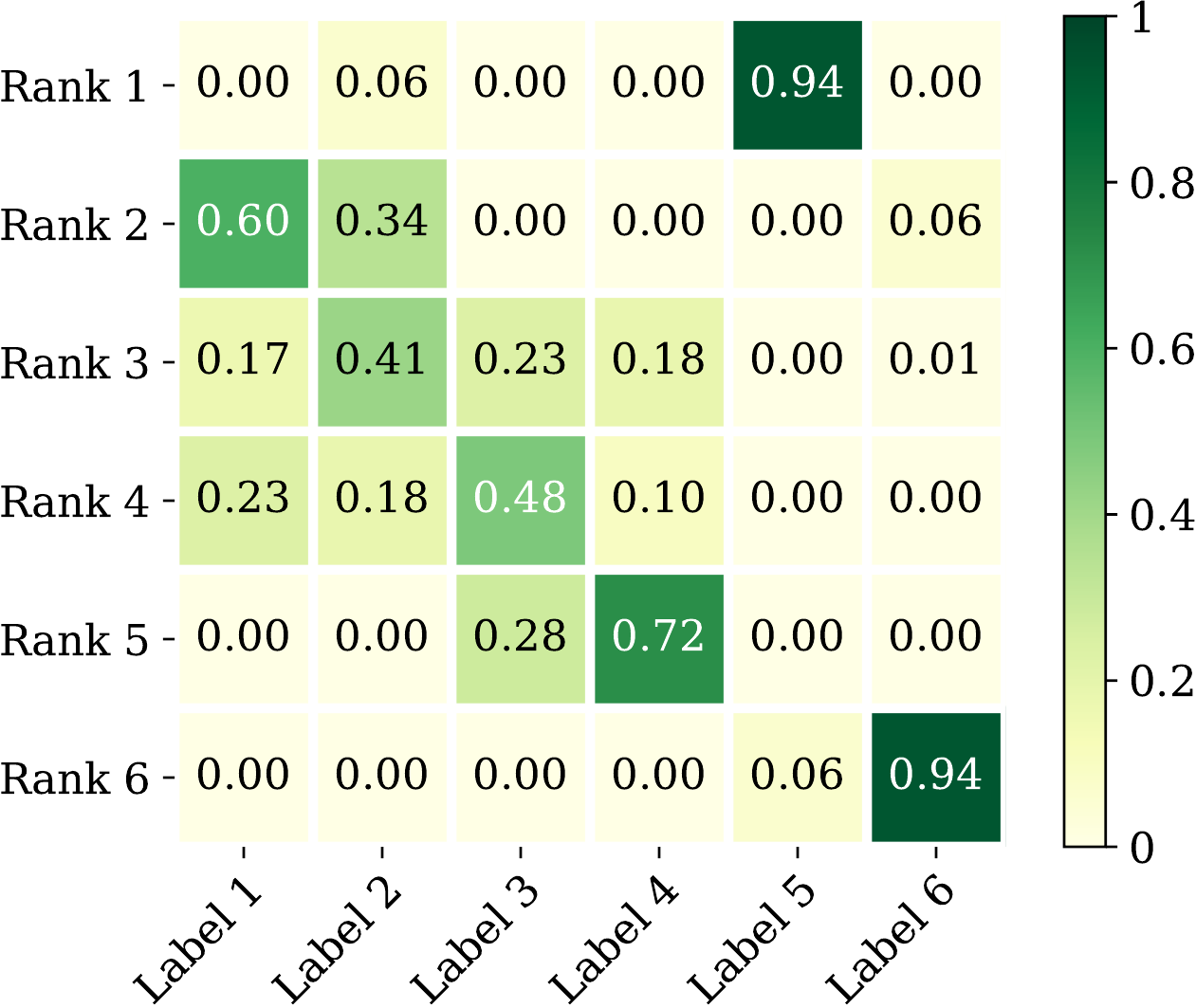}}%
{\caption{\label{fig:soft_permutation_matrix}Example of soft permutation matrix.}}
\end{floatrow}
\end{figure*}

We consider the label ranking setting where supervision is given as
full rankings (e.g., $2 \succ 1 \succ 3 \succ 4$) rather than as label
relevance scores. Note that the exact CRF loss is intractable for this task.
We use the same six public datasets as in
\cite{korba_2018}. We compare different
convex sets for the projection $P_\cC^\Psi$ and 
the decoding $\yhat_L$.  For the Euclidean and KL
projections onto the Birkhoff polytope, we solve the semi-dual formulation
\cite{blondel_2018} by L-BFGS. 
We report the mean Hamming loss, for which our loss is consistent, between the
ground-truth and predicted permutation matrices in the test set. Results are
shown in Table \ref{table:label_ranking_euc} and Table
\ref{table:label_ranking_KL}.
We summarize our findings below.
\begin{itemize}[topsep=0pt,itemsep=2pt,parsep=2pt,leftmargin=10pt]

\item For decoding, using $[0,1]^{k \times k}$ or $\triangle^{k \times k}$
    instead of the Birkhoff polytope considerably degrades accuracy. This
    is not surprising, as these choices do not produce valid permutation
    matrices.

\item Using a squared loss $\frac{1}{2} \|\varphi(y) - \theta\|^2$ ($\cC =
    \RR^{k \times k}$, no projection) works relatively well when combined
    with permutation decoding.
    Using supersets of the Birkhoff polytope as projection set $\cC$, 
    such as $[0,1]^{k \times k}$ or $\triangle^{k \times k}$, 
    improves accuracy substantially. However, the best accuracy is obtained
    when using the Birkhoff polytope for \textbf{both} projections and decoding.

\item The losses derived from Euclidean and KL projections perform similarly.
This is informative, as algorithms for Euclidean projections onto various
sets are more widely available.

\end{itemize}

Beyond accuracy improvements, the projection $\mu = P_\cM^\Psi(W x)$
is useful to visualize soft permutation matrices predicted by the model, an
advantage lost when using supersets of the Birkhoff polytope.

\begin{table}[t]
\caption{Hamming loss (lower is better) for label ranking with Euclidean
projections. The first line indicates 
the projection set $\cC$ used in \eqref{eq:Bregman_projection}. 
The second line indicates the decoding set used in \eqref{eq:calibrated_decoding}.
Using the Birkhoff polytope for \textbf{both} projections and decoding
achieves the best accuracy.
}
\label{table:label_ranking_euc}
\begin{small}
\begin{tabular}{ccccccc}
\toprule
Projection & $[0,1]^{k \times k}$ & $\triangle^{k \times k}$ & $\RR^{k \times
k}$ & $[0,1]^{k \times k}$ & $\triangle^{k \times k}$ & $\cM$ \\
Decoding & $[0,1]^{k \times k}$ & $\triangle^{k \times k}$ & $\cM$ & $\cM$ &
$\cM$ & $\cM$  \\
\midrule
Authorship & 12.83 & 5.62 & 5.70 & 5.18 & 5.70 & \textbf{5.10} \\
Glass & 24.35 & 5.43 & 7.11 & 5.68 & 5.04 & \textbf{4.65} \\
Iris & 27.78 & 10.37 & 19.26 & 4.44 & \textbf{1.48} & 2.96 \\
Vehicle & 26.36 & 7.43 & 9.04 & 7.57 & 6.99 & \textbf{5.88} \\
Vowel & 43.71 & 9.65 & 10.57 & 9.56 & 9.18 & \textbf{8.76} \\
Wine & 10.19 & 1.85 & \textbf{1.23} & 1.85 & 1.85 & 1.85 \\
\bottomrule
\end{tabular}    
\end{small}

{\scriptsize $\cM$: Birkhoff polytope}
\end{table}

\vspace{-0.15cm}
\paragraph{Ordinal regression.} 

We compared classical ridge regression to our order
simplex based loss on sixteen publicly-available datasets
\cite{gutierrez_ordinal_2016}. For evaluation, we use mean absolute error (MAE),
for which our loss is consistent when suitably setting $V$ and $b$
(cf.\ \S\ref{appendix:target_losses}). We find that ridge regression performs the
worst with an average MAE of $0.72$. Combining a squared loss $\frac{1}{2}
\|\varphi(y) - \theta\|^2$ (no projection)
with order simplex decoding at prediction time improves the MAE to $0.47$.  Using a
projection on the unit cube, a superset of the order simplex, further improves
the MAE to $0.45$. Finally, using the Euclidean projection onto the order
simplex achieves the best MAE of $0.43$, confirming that using the order simplex
for both projections and decoding works better. Detailed results are
reported in Table \ref{table:ord_reg}.

\vspace{-0.15cm}
\paragraph{Multilabel classification.}

We compared losses derived from the unit cube and the knapsack polytope
on the same seven datasets as in \cite{sparsemax,fy_losses}.
We set the lower bound $l$ to $0$ and the upper-bound $u$ to 
$\lceil \EE[|Y|] + \sqrt{\mathbb{V}[|Y|]} \rceil$, where $\EE$ and $\mathbb{V}$
are computed over the training set.
Although the unit cube is a strong
baseline, we find that the knapsack polytope improves $F_1$ score
on some datasets, especially with few labels per sample (``birds'',
``emotions'', ``scene''). Results are reported in Tables
\ref{table:multilabel_acc} and \ref{table:multilabel_f1}.

\vspace{-0.35cm}
\section{Conclusion}

We proposed in this paper a general framework for deriving a smooth and convex
loss function from the projection onto a convex set, bringing a computational
geometry perspective to structured prediction. We discussed several examples of
polytopes with efficient Euclidean or KL projection, making our losses useful
for a variety of structured tasks. Our theoretical and empirical results suggest
that the marginal polytope is the convex set of choice when the projection onto
it is affordable.  When not, our framework allows to use any superset with
cheaper-to-compute projection.

\clearpage

\subsubsection*{Acknowledgments}

We thank Vlad Niculae for suggesting the knapsack polytope for multilabel
classification, and Tomoharu Iwata for suggesting to add a lower bound on the
number of labels. We also thank Naoki Marumo for numerous fruitful discussions.


\clearpage
\appendix

\begin{center}
    {\Huge \bf Appendix}
\end{center}

\section{Examples of decomposable target losses}
\label{appendix:target_losses}

For more generality, following \cite{nowak_2019}, we consider losses $L \colon
\cO \times \cY \to \RR_+$, where $\cO$ is the output space and $\cY$ is the
ground-truth space.  Typically, $\cO = \cY$ but we give examples below where
$\cO \neq \cY$.  Our affine decomposition \eqref{eq:loss_decomp} now becomes
\begin{equation}
L(\yhat, y) = \langle \psi(\yhat), V \varphi(y) + b \rangle + c(y),
\label{eq:loss_decomp_general}
\end{equation}
where $\psi \colon \cO \to \RR^{p}$.  
We give examples below of possibly non-convex losses $L$ satisfiying 
decomposition \eqref{eq:loss_decomp_general}.
When not mentioned explicitly, we set $V = I$, $b = \mathbf{0}$ 
and $c(y) = 0$. For more examples of decomposable losses, see also
\cite{ciliberto_2016,nowak_2018,nowak_2019}.

\paragraph{General loss.}

As noted in \cite{ciliberto_2016}, any loss $L \colon \cO \times \cY \to \RR_+$
can always be written as \eqref{eq:loss_decomp_general} if $\cO$ and $\cY$ are
finite sets.  Indeed, it suffices to set $\psi(y) = \varphi(y) = \mathbf{e}_y$
and to define $V$ as the $|\cO| \times |\cY|$ loss matrix, i.e., $V_{\yhat, y} =
L(\yhat, y)$ for all $\yhat \in \cO$ and $y \in \cY$. This, however, ignores
structural information, essential for large output spaces encountered in
structured prediction.

\paragraph{Zero-one loss for multiclass classification.} 

Let $\cO = \cY = [k]$. The 0-1 loss
$L(\yhat, y) = 1[\yhat \neq y]$ can be written as \eqref{eq:loss_decomp_general}
if we set $\psi(y) = \varphi(y) = \mathbf{e}_y$ and $V = 1 - I_{k \times k}$,
i.e., $V$ is the 0-1 cost matrix.

\paragraph{Hamming loss for multilabel classification.}

Let $\cO = \cY = 2^{[k]}$ and $\varphi(y) = \sum_{i=1}^{|y|} \mathbf{e}_{y_i}$.
Then,
\begin{equation}
L(\yhat, y) =  
\sum_{i=1}^k 1[\yhat_i \neq y_i]
= \langle \varphi(\yhat), \mathbf{1} \rangle
+ \langle \varphi(y), \mathbf{1} \rangle
- 2 \langle \varphi(\yhat), \varphi(y) \rangle.
\end{equation}
This can be written as \eqref{eq:loss_decomp_general} with $V = -2 I$, 
$b = \mathbf{1}$ and $c(y) = \langle \varphi(y), \mathbf{1} \rangle$.

\paragraph{Hamming loss for ranking.}

Let $\cO = \cY$ be the set of permutations of $[k]$.
If $\psi(y) = \varphi(y)$ is the permutation matrix associated with permutation
$y$, the Hamming loss is $L(\yhat, y) = \sum_{i=1}^k 1[\yhat_i \neq y_i] = k -
\langle \varphi(\yhat), \varphi(y) \rangle$. It can thus be written as
\eqref{eq:loss_decomp_general} with $V = -I_{k \times k}$ and $c(y) = k$.

\paragraph{Normalized discounted cumulative gain (NDCG).} 

Let $\cO$ be the permutations of $[m]$ and $\cY = [k]^m$ be the
integer relevance scores of $m$ documents. The NDCG loss is defined by
$L(\pi, y) = 1 - \frac{1}{N(y)} \sum_{i=1}^m y_i w_{\pi_i}$, where
$N(y) = \max_{\pi \in \cO} \sum_{i=1}^m y_i w_{\pi_i} = \langle \text{sort}(y),
w \rangle$ is a normalization constant. 
Note that w.l.o.g.\ we use linear gains; exponential gains can be obtained by
$y_i \leftarrow 2^{y_i} - 1$. A typical choice for the discount weights is
$w_i = 1 / \log_2(1 + i)$.
Inspired by \cite{nowak_2019}, we can write $L$ as
\eqref{eq:loss_decomp_general}
by defining $\psi(\pi)$ as the permutation of $w$ according to $\pi$,
$\varphi(y) = y / N(y)$, $V = -I$ and $c(y)=1$. This shows that we can obtain
consistency with respect to the NDCG loss by
regressing \textbf{normalized} relevance scores, $\varphi(y) = y / N(y)$, as
also noted in \cite{ravikumar_2011,nowak_2019}. Furthermore, we note that with
$w_i = 1 / \log_2(1 + i)$, we have $1 = w_1 \ge w_2 \ge \dots \ge w_m \ge 0$,
which implies $\varphi(y) \in [0,1]^m$.  For this reason, we suggest using the
unit cube $[0,1]^m$ for $\cC$ (i.e., the projections are very cheap to compute).
Decoding reduces to linear maximization over the permutahedron, i.e., to sorting
(\S\ref{sec:polytope_examples}).

\paragraph{Precision at $k$ for ranking.}

Let $\cO$ be the permutations of $[m]$ and $\cY = \{0,1\}^m$
be binary relevance scores.
Precision at $k$ corresponds to the number of relevant results
(e.g., labels or documents) in the top $k$ results.  The corresponding loss
can be defined by 
$L(\pi, y) = 1 - \frac{1}{k} \sum_{i=1}^m y_i w_{\pi_i}$, where
$w_1 = \dots = w_k = 1$ and $w_{k+1} = \dots = w_m = 0$
(technically, if the number of positive labels $|y| = \sum_{i=1}^m y_i$ is less
than $k$, we replace $k$ with $|y|$).
Similarly as for NDCG, we can therefore write $L$ as
\eqref{eq:loss_decomp_general}, which suggests that regressing $\varphi(y) = y /
k$ is consistent with precision at $k$.
Again, decoding reduces to linear maximization over the permutahedron.

\paragraph{Absolute loss for ordinal regression.}

Let $\cO = \cY = [k]$ and $\psi(y) = \varphi(y) = \sum_{i < y} \mathbf{e}_i$.
Then,
\begin{equation}
L(\yhat, y) 
= |\yhat - y| 
= \langle \varphi(\yhat), \mathbf{1} \rangle
+ \langle \varphi(y), \mathbf{1} \rangle
- 2 \langle \varphi(\yhat), \varphi(y) \rangle.
\end{equation}
This can be written as \eqref{eq:loss_decomp_general} with $V = -2 I$, 
$b = \mathbf{1}$ and $c(y) = \langle \varphi(y), \mathbf{1} \rangle$.
A similar loss decomposition is derived in \cite{nowak_2019} in the case of a
signed encoding, instead of the zero-one encoding we use.
However, the signed encoding is problematic when using KL projections and does
not lead to sparse projections when using Euclidean projections.

\section{Experiment details and additional empirical results}

We discuss in this section our experimental setup and additional empirical
results.

For all datasets, we normalized samples to have zero mean unit variance.  We use
the train-test split from the dataset when provided. When not, we use 80\% for
training data and 20\% for test data. We hold out 25\% of the training data for
hyperparameter validation purposes. 
For the regularization hyper-parameter $\lambda$, we used ten log-spaced values
between $10^{-4}$ and $10^{4}$.
Once we select the best hyperparameter, we
refit the model on the entire training set.
We ran all experiments on a machine with Intel(R) Xeon(R) CPU with 2.90GHz
and 4GB of RAM.

In all experiments, we use publicly-available datasets:
\begin{itemize}
\item \url{https://github.com/akorba/Structured_Approach_Label_Ranking}
\item \url{http://www.uco.es/grupos/ayrna/ucobigfiles/datasets-orreview.zip}
\item \url{http://mulan.sourceforge.net/datasets-mlc.html}
\item \url{https://www.csie.ntu.edu.tw/~cjlin/libsvmtools/datasets/}
\end{itemize}

\subsection{Label ranking}

In this section, we compare the Birkhoff and permutahedron
polytopes for the same label ranking task as described in the main manuscript.
With the Birkhoff polytope, $\theta = g(x) \in \RR^{k \times k}$ can be
interpreted as an affinity matrix between classes for the input $x$. With the
permutahedron, $\theta = g(x) \in \RR^k$ can be intepreted as vector, containing
the score of each class for input $x$.
Therefore, the two polytopes have different expressive power. For the model
$g(x)$, we compare $g(x) = Wx$, where $W$ is a matrix or linear
map of proper shape, and a polynomial model $g(x) = \sum_{i=1}^n w_i
\kappa(x, x_i)$, where $w \in \RR^n$, $\kappa(x, x') \coloneqq (\langle x, x'
\rangle + 1)^D$ and $D$ is the polynomial degree.
For the Euclidean projection
onto the permutahedron, we use the isotonic regression solver from
scikit-learn \cite{scikit_learn}. 

Our results, shown in Table \ref{table:birkhoff_permutahedron_cmp}, indicate that in
the case of a linear model, the Birkhoff polytope outperforms the permutahedron
by a large margin. Using a polynomial model closes the gap between the two, but
the model based on the Birkhoff polytope is still slightly better.

\begin{table}[ht]
\caption{Test Hamming loss comparison when using the Birkhoff and permutahedron
polytopes.}
\label{table:birkhoff_permutahedron_cmp}
\begin{tabular}{ccccc}
\toprule
Projection & $\cB$ & $\cP$ & $\cP$ & $\cP$ \\
Decoding & $\cB$ & $\cP$ & $\cP$ & $\cP$ \\
Model & Linear & Linear & Poly ($D=2$) & Poly ($D=3$) \\
\midrule
Authorship & \textbf{5.10} & 10.06 & 10.50 & 8.59 \\
Glass & \textbf{4.65} & 7.49 & 7.10 & 8.14 \\
Iris & \textbf{2.96} & 27.41 & 20.00 & 5.93 \\
Vehicle & \textbf{5.88} & 11.62 & 8.30 & 9.26 \\
Vowel & \textbf{8.76} & 14.35 & 11.74 & 10.21 \\
Wine & \textbf{1.85} & 8.02 & 3.08 & 6.79 \\
\bottomrule
\end{tabular}    

{\scriptsize $\mathcal{B}$: Birkhoff polytope, $\cP$: permutahedron}
\end{table}

\newpage
\subsection{Ordinal regression}

In this section, we present our detailed results on ordinal regression.
Table \ref{table:ord_reg} below shows the results for each dataset.
For context, the first column indicates a simple baseline in which we always
predict the median label calculated on the train set. The second column
indicates classical ridge regression where we used rounding to the closest
integer as decoding.  Using the order simplex for both projections and decoding
achieves the best MAE on average.

\begin{table}[ht]
\caption{Mean absolute error (MAE) of our losses with Euclidean projections.}
\label{table:ord_reg}
\begin{tabular}{cccccc}
\toprule
Projection & \multirow{2}{*}{Baseline} & $\RR$ & $\RR^{k-1}$ & $[0,1]^{k-1}$ &
$\mathcal{M}$ \\
Decoding & & Round & $\mathcal{M}$ & $\mathcal{M}$ & $\mathcal{M}$ \\
\midrule
ERA & 1.61 & 1.24 & \textbf{1.19} & \textbf{1.19} & \textbf{1.19} \\
ESL & 1.12 & 0.34 & 0.37 & \textbf{0.30} & \textbf{0.30} \\
LEV & 0.71 & \textbf{0.41} & 0.48 & 0.46 & 0.46 \\
SWD & 0.63 & 0.47 & \textbf{0.41} & 0.42 & 0.42 \\
Automobile & 1.02 & 0.50 & 0.54 & 0.54 & \textbf{0.48} \\
Balance-scale & 0.91 & 0.29 & \textbf{0.10} & \textbf{0.10} & \textbf{0.10} \\
Car & 0.41 & 0.32 & 0.23 & \textbf{0.17} & \textbf{0.17} \\
Contact-lenses & 0.5 & 0.50 & 0.33 & \textbf{0.17} & \textbf{0.17} \\
Eucalyptus & 1.22 & 3.96 & \textbf{0.40} & 0.44 & 0.45 \\
Newthyroid & 0.29 & 0.09 & 0.11 & \textbf{0.02} & \textbf{0.02} \\
Pasture & 0.66 & 0.44 & \textbf{0.33} & \textbf{0.33} & \textbf{0.33} \\
Squash-stored & 0.54 & \textbf{0.38} & 0.54 & 0.62 & 0.62 \\
Squash-unstored & 0.54 & 0.46 & 0.31 & 0.31 & \textbf{0.15} \\
Tae & 0.69 & \textbf{0.66} & \textbf{0.66} & \textbf{0.66} & \textbf{0.66} \\
Toy & 0.97 & \textbf{0.96} & 0.97 & 0.97 & 0.97 \\
Winequality-red & 0.66 & 0.46 & \textbf{0.45} & 0.46 & 0.46 \\
\midrule
Average MAE & 0.78 & 0.72 & 0.47 & 0.45 & 0.43 \\
\midrule
Average rank & 4.75 & 2.9 & 2.1 & 1.6 & \textbf{1.5} \\
\bottomrule
\end{tabular}    

{\scriptsize $\mathcal{M}$: order simplex}
\end{table}

\newpage
\subsection{Multilabel classification}

In this section, we show full empirical results for our multilabel experiment.
Dataset statistics are summarized in Table \ref{table:multilabel_datasets}.
Empirical results are shown in Tables \ref{table:multilabel_acc} and 
\ref{table:multilabel_f1}.

\begin{table}[ht]
    \caption{Multilabel dataset statistics}
    \label{table:multilabel_datasets}
    \centering
    \begin{tabular}{r c c c c c c c}
        \toprule
        Dataset & Type & Train & Dev & Test & Features &
        Classes & Avg. labels \\
        \midrule
        Birds & Audio & 134 & 45 & 172 & 260 & 19 & 1.96 \\
        Cal500 & Music & 376 & 126 & 101 & 68 & 174 & 25.98 \\
        Emotions & Music & 293 & 98 & 202 & 72 & 6 & 1.82 \\
        Mediamill & Video & 22,353 & 7,451 & 12,373 & 120 & 101 & 4.54 \\
        Scene & Images & 908 & 303 & 1,196 & 294 & 6 & 1.06\\
        SIAM TMC & Text & 16,139 & 5,380 & 7,077 & 30,438 & 22 & 2.22\\
        Yeast & Micro-array & 1,125 & 375 & 917 & 103 & 14 & 4.17\\
        \bottomrule
    \end{tabular}
\end{table}

\begin{table}[ht]
\caption{Accuracy comparison on multilabel classification, using Euclidean
projections.}
\label{table:multilabel_acc}
\begin{tabular}{ccccc}
\toprule
Projection & $[0,1]^k$ & $\RR^k$ & $[0,1]^k$ & $\cM$ \\
Decoding & $[0,1]^k$ & $\cM$ & $\cM$ & $\cM$ \\
\midrule
Birds & 90.21 & 90.61 & 90.61 & \textbf{90.70} \\
Cal500 & 85.78 & 85.76 & 85.78 & \textbf{85.80} \\
Emotions & \textbf{77.64} & 76.32 & 75.99 & 75.83 \\
Mediamill & \textbf{96.90} & 96.85 & \textbf{96.90} & 96.82 \\
Scene & 90.05 & 88.34 & 89.62 & \textbf{90.79} \\
TMC & \textbf{94.67} & 94.48 & 94.65 & \textbf{94.67} \\
Yeast & \textbf{79.75} & 79.70 & \textbf{79.75} & 79.71 \\
\bottomrule
\end{tabular}    

{\scriptsize $\mathcal{M}$: knapsack polytope}
\end{table}

\begin{table}[ht]
\caption{$F_1$-score comparison on multilabel classification, using Euclidean
projections.}
\label{table:multilabel_f1}
\begin{tabular}{ccccc}
\toprule
Projection & $[0,1]^k$ & $\RR^k$ & $[0,1]^k$ & $\cM$ \\
Decoding & $[0,1]^k$ & $\cM$ & $\cM$ & $\cM$ \\
\midrule
Birds & 38.87 & 37.75 & 39.21 & \textbf{39.43} \\
Cal500 & 34.62 & \textbf{35.86} & 34.63 & 34.61 \\
Emotions & 56.60 & 51.73 & 53.98 & \textbf{62.57} \\
Mediamill & \textbf{56.22} & 55.35 & \textbf{56.22} & 54.53 \\
Scene & 61.06 & 50.33 & 58.95 & \textbf{69.01} \\
TMC & \textbf{60.45} & 58.61 & 60.37 & 60.25 \\
Yeast & \textbf{60.24} & 60.20 & 60.23 & 60.06 \\
\bottomrule
\end{tabular}    

{\scriptsize $\mathcal{M}$: knapsack polytope}
\end{table}

\clearpage
\section{Proofs}

\subsection{Proof of strong convexity of Shannon negentropy (Proposition \ref{prop:strong_convexity})}
\label{appendix:proof_prop_strong_convexity}

Let $\cC \subseteq \RR_+^d$ and $\Psi(u) = \langle u, \log u \rangle$.
For all $u, v \in \cC$ we have \cite[\S 9.1.2]{boyd_2004}
\begin{equation}
    \Psi(u) = \Psi(v) + \nabla \Psi(v)^\top (u - v)
    + \frac{1}{2} (u - v)^\top \nabla^2 \Psi(w) (u - v),
\end{equation}
for some $w \in \cC$ in the line segment $[u,v]$, and
where $\nabla \Psi(u) = \log u + 1$, $\nabla^2 \Psi(u) =
\text{diag}(u^{-1})$.
Recall that $\Psi$ is $\frac{1}{\beta}$-strongly convex over $\cC$ 
w.r.t. $\|\cdot\|$ if for all $u,v \in \cC$
\begin{equation}
    \Psi(u) \ge \Psi(v) + \nabla \Psi(v)^\top (u - v)
    + \frac{1}{2\beta} \|u - v\|^2.
\label{eq:strongly_convex}
\end{equation}
Therefore, letting $z=u-v$, it suffices to show that for all $u, v, w \in \cC$
\begin{equation}
\beta z^\top \nabla^2 \Psi(w) z \ge \|z\|_1^2.
\label{eq:Hessian_inequality}
\end{equation}
Note that if there exists $w_i = 0$, then \eqref{eq:Hessian_inequality} clearly
holds. Therefore we can focus on showing \eqref{eq:Hessian_inequality} for $w > 0$.
This is indeed verified since by the Cauchy–Schwarz inequality 
\begin{equation}
\|z||_1^2 
= \left(\sum_{i=1}^d \frac{|z_i|}{\sqrt{w_i}} \sqrt{w_i}\right)^2
\le \sum_{i=1}^d \frac{z_i^2}{w_i} \sum_{i=1}^d w_i
= z^\top \nabla^2 \Psi(w) z ~ ||w||_1.
\end{equation}
Therefore $\Psi$ is $\frac{1}{\beta}$-strongly convex over $\cC$ w.r.t. $\|\cdot\|_1$, 
with $\beta = \sup_{w' \in \cC} ||w'||_1$.


\subsection{Projection onto the knapsack polytope (Proposition
\ref{prop:projection_budget})}
\label{appendix:proj_budget}

\paragraph{Euclidean case.}

The Euclidean projection onto the knapsack polytope is
\begin{equation}
    \argmin_{\mu \in \RR^k} \frac{1}{2} \|\mu - \theta\|^2
    \quad \text{s.t.} \quad
    l \le \langle \mu, \ones \rangle \le u, \quad 0 \le \mu \le 1.
\end{equation}
The corresponding Lagrangian is
\begin{equation}
    \mathcal{L} = \frac{1}{2} \|\mu - \theta\|^2 
    + \tau (\langle \mu, \ones \rangle - u) + \eta (l - \langle \mu, \ones \rangle)
    - \langle \xi, \mu \rangle + \langle \zeta, \mu - \ones \rangle,
\end{equation}
where the dual feasibility conditions are
$\tau \ge 0$, $\eta \ge 0$, $\xi \in \RR_+^k$ and $\zeta \in \RR_+^k$.
From the stationary conditions, the optimal $\mu$ should satisfy
\begin{equation}
    \mu = \theta + (\eta - \tau) \ones + \xi - \zeta.
\end{equation}
From the complementary slackness conditions,
\begin{align}
\tau (\langle \mu, \ones \rangle - u) &= 0 \\
\eta (l - \langle \mu, \ones \rangle) &= 0 \\
\zeta_i (\mu_i - 1) &= 0 \quad \forall i \in [k] \\
\xi_i \mu_i &= 0 \quad \forall i \in [k].
\end{align}

If $0 < \mu_i < 1$, then $\xi_i = \zeta_i = 0$ and
$\mu_i = \theta_i - \tau + \eta$.
If $\xi_i > 0$ then $\mu_i = 0$. If $\zeta_i > 0$ then $\mu_i = 1$.
Altogether, we thus have for all $i \in [k]$
\begin{equation}
    \mu_i = \text{clip}_{[0,1]}(\theta_i - \tau + \eta)
    \coloneqq
    \min\{1, \max\{0, \theta_i - \tau + \eta\} \}.
\end{equation}
Three cases can happen:
\begin{itemize}
    \item If $\tau = \eta = 0$, the inequality $l \le \langle \mu, \ones \rangle
        \le u$ is inactive. Therefore the projection
        $\text{clip}_{[0,1]}(\theta)$ onto the unit cube is optimal.

\item If $\tau > 0$, then $\eta=0$ and $\langle \mu, \ones \rangle = u$ is
    active.  This case happens when $\langle \text{clip}_{[0,1]}(\theta), \ones
    \rangle > u$.
 
\item If $\eta > 0$, then $\tau=0$ and $l = \langle \mu, \ones \rangle$ is
    active.  This case happens when $\langle \text{clip}_{[0,1]}(\theta), \ones
    \rangle < l$.

\end{itemize}
The second and third cases correspond to a projection onto
$\{ \mu \in \RR^k \colon \langle \mu, \ones \rangle = m, 0 \le \mu \le 1 \}$,
with $m = u$ or $m = l$. This projection can be computed in $O(k)$ time
using Pardalos and Kovoor's algorithm \cite{pardalos_1990}.
See also \cite[Appendix A]{almeida_2013} for pseudo code.
Since that set is a special case of permutahedron with
$w \in \RR^k$ defined by $w_1 = \dots = w_m = 1$ and $w_{m+1} = \dots = w_k =
0$, we can also use the projection onto the permutahedron. 
The cost is $O(k \log k)$ for sorting $\theta$ and $O(k)$ for isotonic
regression via the pool adjacent violators algorithm
\cite{projection_permutahedron}.
Yet another alternative is to search for $\tau$ solving
$\sum_{i=1}^k \text{clip}_{[0,1]}(\theta_i - \tau) = m$ by bisection.

\paragraph{KL case.} We want to solve
(note that the non-negativity constraint on $\mu$ is vacuous)
\begin{equation}
    \argmin_{\mu \in \RR^k} 
    \langle \mu, \log \mu \rangle - \langle \mu, \theta \rangle
    \quad \text{s.t.} \quad
    l \le \langle \mu, \ones \rangle \le u, \quad 0 \le \mu \le 1.
\end{equation}
As for the Euclidean projection, we consider three cases.
\begin{itemize}
    \item If the projection $\nu = \min(1, e^{\theta - 1})$ on the unit cube
        satisfies the constraints, $\nu$ is the optimal solution.

    \item If $\langle \nu, \ones \rangle > u$, we need to satisfy the
        constraint $\langle \mu, \ones \rangle = u$.

    \item If $\langle \nu, \ones \rangle < l$, we need to satisfy the
        constraint $\langle \mu, \ones \rangle = l$.
\end{itemize}
The last two cases correspond to solving the problem
\begin{equation}
    \argmin_{\mu \in \RR^k} 
    \langle \mu, \log \mu \rangle - \langle \mu, \theta \rangle
    \quad \text{s.t.} \quad
    \langle \mu, \ones \rangle = m, \quad \mu \le 1.
\end{equation}
with $m=u$ or $m=l$. We can rewrite it as
\begin{equation}
    \argmin_{\alpha \in \RR^k} 
    \langle \alpha, \log \alpha \rangle - \langle \alpha, z \rangle
    \quad \text{s.t.} \quad
    \langle \alpha, \ones \rangle = 1, \quad \alpha \le \frac{1}{m}.
\end{equation}
with $\alpha \coloneqq \frac{\mu}{m}$ and $z \coloneqq \theta - (\log m) \ones$.
An $O(k \log k)$ algorithm for solving this constrained softmax (KL projection
onto a capped simplex) was derived in \cite{easy_first}.
A related projection using a different entropy is derived in \cite{amos_2019}.

\subsection{Vertices of the order simplex (Proposition \ref{prop:order_simplex})}
\label{appendix:vertices_order_simplex}

Let us gather the vertices $\varphi(y) \in \{0,1\}^{k-1}$ for all $y \in [k]$ as
columns in a matrix $M \in \{0,1\}^{k-1 \times k}$.  For instance, with $k=4$,
\begin{equation}
M =
\begin{bmatrix}
    0 & 1 & 1 & 1 \\
    0 & 0 & 1 & 1 \\
    0 & 0 & 0 & 1
\end{bmatrix}.
\end{equation}
Recall that
\begin{equation}
\cM = M \triangle^k
= \{M p \colon p \in \RR^k, p \ge 0, \langle p, \ones \rangle = 1\} \subset \RR^{k-1}.
\end{equation}
Let $\mu = Mp$. Then for all $p \in \triangle^k$
\begin{align}
\mu_1 &=  p_2 + p_3 + \dots + p_k \\
\mu_2 &=  p_3 + p_4 + \dots + p_k \\
\mu_3 &=  p_4 + \dots + p_k \\
 & \vdots \\
\mu_{k-1} &= p_k,
\end{align}
from which we obtain
\begin{align}
    1 - \mu_1 &= p_1 \ge 0 \\
    \mu_1 - \mu_2 &= p_2  \ge 0\\
    \mu_2 - \mu_3 &= p_3  \ge 0\\
              & \vdots \\
    \mu_{k-1} &= p_k \ge 0.
\end{align}
Notice that $\langle p, \ones \rangle = 1$ is automatically satisfied for any
$\mu = Mp$.  Therefore 
\begin{equation}
    \cM = \{\mu \in \RR^{k-1} \colon 1 \ge \mu_1 \ge \mu_2 \ge \dots \ge \mu_{k-1} \ge 0 \},
\end{equation}
which is known as the order simplex \cite{grotzinger_1984}.

\subsection{Calibration of target and surrogate excess risks (Proposition
\ref{prop:calibration})}
\label{appendix:proof_prop_calibration}

We prove Proposition \ref{prop:calibration}, extending a recent analysis
\cite{nowak_2019} to the more general projection losses.

\subsubsection{Background}

In this section, after reviewing the classical notions of pointwise
and population excess risks, we
discuss calibration functions for structured prediction, as introduced in
\cite{osokin_2017}. We use the generalized notation introduced in
\S\ref{appendix:target_losses}, with output space $\cO$ and ground truth space
$\cY$.

\paragraph{Pointwise and population risks.}

Given a distribution $q \in \triangle^{|\cY|}$, we define
the \emph{pointwise} risk of $\yhat \in \cO$ for the loss $L$ and the pointwise
risk of $\theta \in \Theta$ for the surrogate $S$ by
\begin{equation}
    \ell(\yhat, q) \coloneqq \EE_{Y \sim q} ~ L(\yhat, Y)
    \quad \text{and} \quad
    s(\theta, q) \coloneqq \EE_{Y \sim q} ~ S(\theta, Y),
\end{equation}
respectively. We also define the corresponding excess of pointwise risks,
the difference between the pointwise risks and the pointwise Bayes risk:
\begin{equation}
\delta \ell(\yhat, q) \coloneqq \ell(\yhat, q) - \inf_{y' \in \cO} \ell(y', q)
\quad \text{and} \quad
\delta s(\theta, q) \coloneqq s(\theta, q) - \inf_{\theta' \in \Theta}
s(\theta', q).
\label{eq:pointwise_risk}
\end{equation}

Given a joint distribution $p \in \triangle(\cX \times \cY)$,
let us now define the \emph{population} target risk 
and the population surrogate risk by
\begin{equation}
\cL(f) \coloneqq \EE_{(X,Y) \sim p} ~ L(f(X), Y)
\quad \text{and} \quad
\cS(g) \coloneqq \EE_{(X,Y) \sim p} ~ S(g(X), Y).
\end{equation}
The quality of estimators $f$ and $g$ is measured in terms of the excess of
population risks
\begin{equation}
\delta \cL(f) \coloneqq \cL(f) - \inf_{f'\colon \cX \to \cO} \cL(f')
\quad \text{and} \quad
\delta \cS(g) \coloneqq \cS(g) - \inf_{g' \colon \cX \to \Theta} \cS(g').
\end{equation}
Note that the population risks can be written in terms of the pointwise ones as
\begin{equation}
\cL(f) = \EE_{X \sim p_{\cX}} ~ \ell(f(X), p(\cdot | X))
\quad \text{and} \quad
\cS(g) = \EE_{X \sim p_{\cX}} ~ s(g(X), p(\cdot | X)),
\label{eq:population_using_pointwise}
\end{equation}
where $p(\cdot|x)$ is the conditional distribution over $\cY$,
and $p_{\cX}$ is the marginal distribution over $\cX$. 
Analogously, when the surrogate is a F-Y loss generated by $\Omega$, we will use $s_\Omega$,
$\delta s_\Omega$, $\cS_\Omega$ and $\delta \cS_\Omega$.

\paragraph{Calibration functions.}

Let $d \colon \Theta \to \cO$ be a decoding function, namely a function that
turns a continuous prediction $\theta = g(x)$ into a discrete structure in
$\cO$.
A calibration function $\zeta$ is a function relating the excess 
of \emph{pointwise} risks $\delta \ell$ and $\delta s$ 
for all $\theta \in \Theta$ and $q \in \triangle^{|\cY|}$ by
\begin{equation}
    \zeta(\delta \ell(d(\theta), q)) \le \delta s(\theta, q).
\end{equation}
It allows to control how much reduction of $\delta s$ is needed to
reduce $\delta \ell$ when using $d$ as a decoder (larger $\zeta$ is better).
As shown in \cite{osokin_2017}, $\zeta$ can be cast as an optimization problem,
\begin{equation}
\zeta(\epsilon) =
\inf_{\theta \in \Theta, q \in \triangle^{|\cY|}} \delta s(\theta, q) 
\quad \text{s.t.} \quad
\delta \ell(d(\theta), q) \ge \epsilon.
\label{eq:calibration_function}
\end{equation}
It is easy to verify that $\zeta$ is positive, non-decreasing, and satisfies
$\zeta(0)=0$. As shown in \cite{osokin_2017,nowak_2019}, any convex lower-bound
$\xi$ of $\zeta$ allows to in turn calibrate the excess of \emph{population} risks:
\begin{equation}
    \xi(\delta \cL(d \circ g)) \le \delta \cS(g),
    \label{eq:population_risk_calibration}
\end{equation}
for all $g \colon \cX \to \Theta$ and $d \colon \Theta \to \cO$.
This follows from Jensen's inequality and from
\eqref{eq:population_using_pointwise}.
If $\xi(\epsilon) > 0$ for all $\epsilon > 0$ and $\xi(0) = 0$, this implies
Fisher consistency.

\subsubsection{Calibration function of Fenchel-Young losses}

We derive the exact expression of the calibration function
\eqref{eq:calibration_function} for \textbf{general} Fenchel-Young losses.
\begin{lemma}{Calibration function of general Fenchel-Young losses}
   
Let $L(\yhat, y)$ be decomposed as \eqref{eq:loss_decomp_general}
and $S(\theta, y) = S_\Omega(\theta, \varphi(y))
\coloneqq \Omega^*(\theta) + \Omega(\varphi(y)) - \langle \theta, \varphi(y) \rangle$,
with $\varphi(\cY) \subseteq \dom(\Omega)$.  
Then the calibration function \eqref{eq:calibration_function}
with decoder $d \colon \Theta \to \cO$ reads
\begin{equation}
\zeta(\epsilon) =
\inf_{\theta \in \Theta, \mu \in \cM}
S_\Omega(\theta, \mu)
\quad \text{s.t.} \quad
\langle \psi(d(\theta)) - \psi(\yhat_L(\mu)), V \mu + b \rangle
\ge \epsilon.
\end{equation}
\end{lemma}
\textbf{Proof.} Let $\muv(q) \coloneqq \EE_{Y \sim q}[\varphi(Y)]$.
The pointwise surrogate risk reads
\begin{align}
s_\Omega(\theta, q) 
&\coloneqq  \EE_{Y \sim q}[S_\Omega(\theta, \varphi(Y))] \\
&= \sum_{y \in \cY} q(y) ( \Omega^*(\theta) + \Omega(\varphi(y)) - 
\langle \theta, \varphi(y) \rangle) \\
&= \Omega^*(\theta) + \Omega(\muv(q)) - \langle \theta, \muv(q) \rangle
+ \EE_{Y \sim q}[\Omega(\varphi(Y))] - \Omega(\muv(q)) \\
&\eqqcolon S_\Omega(\theta, \muv(q)) + I_\Omega(\varphi(Y), q),
\end{align}
where we defined $I_\Omega(\varphi(Y), q)$, the Bregman information
\cite{bregman_clustering} of
the random variable $\varphi(Y)$ with generating function $\Omega$.
Hence the excess of pointwise surrogate risk reads
\begin{align}
\delta s_\Omega(\theta, q) 
&= s_\Omega(\theta, q) - \inf_{\theta' \in \Theta} s_\Omega(\theta', q) \\
&= S_\Omega(\theta, \muv(q)) - 
\underbrace{\inf_{\theta' \in \Theta} S_\Omega(\theta', \muv(q))}_{= ~ 0} \\
&= S_\Omega(\theta, \muv(q)),
\end{align}
where in the second line we used \cite[Proposition 2]{fy_losses}.
The pointwise target risk reads
\begin{align}
\ell(\yhat, q) 
&= \EE_{Y \sim q}[L(\yhat, Y)] \\
&= \sum_{y \in \cY} q(y) (\langle \psi(\yhat), V \varphi(y) + b \rangle + c(y)) \\
&= \langle \psi(\yhat), V \muv(q) + b \rangle + \EE_{Y \sim q}[c(Y)].
\end{align}
Hence the excess of pointwise target risk reads
\begin{align}
\delta \ell(\yhat, q)  
&= \ell(\yhat, q) - \inf_{y' \in \cO} \ell(y', q) \\
&= \langle \psi(\yhat), V \muv(q) + b \rangle
- \inf_{y' \in \cO} \langle \psi(y'), V \muv(q) + b \rangle \\
&= \langle \psi(\yhat), V \muv(q) + b \rangle
- \langle \psi(\yhat_L(\muv(q))), V \muv(q) + b \rangle \\
&= \langle \psi(\yhat) - \psi(\yhat_L(\muv(q))), V \muv(q) + b \rangle,
\end{align}
where
\begin{equation}
\yhat_L(u) 
\coloneqq \argmin_{y' \in \cO} \langle \psi(y'), V u + b \rangle.
\end{equation}
Therefore we can rewrite \eqref{eq:calibration_function} as
\begin{equation}
\zeta(\epsilon) =
\inf_{\theta \in \Theta, q \in \triangle^{|\cY|}}
S_\Omega(\theta, \muv(q))
\quad \text{s.t.} \quad
\langle \psi(d(\theta)) - \psi(\yhat_L(\muv(q))), V \muv(q) + b \rangle
\ge \epsilon.
\end{equation}
Using the change of variable $\mu = \muv(q) \in \cM = \conv(\varphi(\cY))$
gives the desired result. $\square$

\subsubsection{Technical lemma}

We give in this section a technical lemma, which will be useful for the rest of
the proof.

\begin{lemma}{Upper-bound on pointwise target risk}
\label{lemma:technical_lemma}

Let $\sigma \coloneqq \sup_{\yhat \in \cO} \|V^\top \psi(\yhat)\|_*$, where
$\|\cdot\|_*$ denotes the dual norm of $\|\cdot\|$. Then,
\begin{equation}
\delta \ell(\yhat_L(u), q) \le 2 \sigma \|\muv(q) - u \|
\quad \forall u \in \RR^p, q \in \triangle^{|\cY|}.
\end{equation}
\end{lemma}

\textbf{Proof.} The proof is a slight modification of
\cite[Lemma D.3]{nowak_2019} and is included for completeness. In that work, $V
= I$ and $b=0$, or put differently, they are absorbed into $\varphi$. In this
work, we keep $V$ and $b$
explicitly to decouple the label encoding from the loss decomposition. This is
important in order to keep MAP and projection algorithms unchanged.
Following \cite{nowak_2019}, let us decompose 
$\delta \ell (\yhat_L(u), q)$ into two terms $A$ and $B$:
\begin{equation}
\begin{aligned}
\delta \ell (\yhat_L(u), q) 
&= \langle \psi(\yhat_L(u)) - \psi(\yhat_L(\muv(q))), V \muv(q) + b \rangle \\
&= \underbrace{\langle \psi(\yhat_L(u)), V \muv(q) + b - V u - b \rangle}_A
+ \underbrace{\langle \psi(\yhat_L(u)), V u + b \rangle 
- \langle \psi(\yhat_L(\muv(q)), V \muv(q) + b \rangle}_B.
\end{aligned}
\end{equation}
Clearly, $A \le \sup_{\yhat \in \cO} ~ | \langle \psi(\yhat), V \muv(q) - V u
\rangle |$.

Using $|\min_z \eta(z) - \min_z \eta'(z)| \le \sup_z |\eta(z) - \eta'(z)|$,
an inequality also used in \cite[Thm. 12]{ciliberto_2016},
we also get
$B \le \sup_{\yhat \in \cO} ~ | \langle \psi(\yhat), V \muv(q) - V u \rangle |$.

Therefore, in both cases, we see that $b$ cancels out.
Combining the two, we obtain
\begin{equation}
\delta \ell(\yhat_L(u), q) 
\le 2 \sup_{\yhat \in \cO} ~ | \langle \psi(\yhat), V \muv(q) - V u \rangle|
= 2 \sup_{\yhat \in \cO} ~ | \langle V^\top \psi(\yhat), \muv(q) - u \rangle|.
\end{equation}
By definition of the dual norm, we have the (generalized) Cauchy-Shwarz
inequality $\langle x, y \rangle \le \|x\| ~ \|y\|_*$. Therefore,
\begin{equation}
\delta \ell(\yhat_L(u), q) 
\le 
2 \sup_{\yhat \in \cO} |\langle V^\top \psi(\yhat), \muv(q) - u \rangle| 
\le
2 \sup_{\yhat \in \cO} \|V^\top \psi(\yhat)\|_* \|\muv(q) - u \|. 
\end{equation}
\hfill $\square$

\subsubsection{Convex lower bound on the calibration function of
projection-based losses}

Since $\zeta$ above could be non-convex and difficult to compute,
we next derive a convex lower bound for a particular \textbf{subset} of
Fenchel-Young losses (namely, projection-based losses $S_\cC^\Psi$) and for a
particular decoder $d \colon \Theta \to \cO$ (namely, $d = \yhat_L \circ
P_\cC^\Psi$).
\begin{lemma}{Convex lower bound on the calibration function of
$S_\cC^\Psi$}
\label{lemma:lower_bound}

Let $L(\yhat, y)$ be decomposed as \eqref{eq:loss_decomp_general} and $S(\theta, y) =
S_\cC^\Psi(\theta, y)$ be defined as in \eqref{eq:loss_shorthand}.
Assume $\Psi$ is $\frac{1}{\beta}$-strongly convex over $\cC$ w.r.t.
$\|\cdot\|$, Legendre-type,
and $\cC$ is a convex set such that $\varphi(\cY) \subseteq \cC \subseteq
\dom(\Psi)$.  
Let $\sigma \coloneqq \sup_{\yhat \in \cO} \|V^\top \psi(\yhat)\|_*$,
where $\|\cdot\|_*$ is the dual norm of $\|\cdot\|$. Then, the calibration
function defined \eqref{eq:calibrated_decoding} 
with $d = \yhat_L \circ P_\cC^\Psi$
is lower bounded as 
\begin{equation}
\zeta(\epsilon) \ge \frac{\epsilon^2}{8 \beta \sigma^2}.
\end{equation}
\end{lemma}

\textbf{Proof.}
Let us set $\Omega = \Psi + I_\cC$, where $\Psi$ is Legendre-type.
Note that this does not imply that $\Omega$ itself is Legendre-type.
Using Lemma \ref{lemma:technical_lemma}, we have for all $u \in \RR^p$ and all
$q \in \triangle^{|\cY|}$
\begin{equation}
\delta \ell(\yhat_L(u), q) 
\le 2 \sigma \|\muv(q) - u \|.
\end{equation}
Let us set the decoder to
$d = \yhat_L \circ \nabla \Omega^*$.
With $u = \nabla \Omega^*(\theta)$, we thus get
for all $\theta \in \Theta$ and $q \in \triangle^{|\cY|}$:
\begin{equation}
\epsilon
\le
\delta \ell(d(\theta), q) 
\le 
2 \sigma \|\muv(q) - \nabla \Omega^*(\theta)\|.
\end{equation}
From \eqref{eq:strongly_convex}, $\Psi$ is $\frac{1}{\beta}$-strongly convex
over $\cC$ w.r.t.  $\|\cdot\|$ if and only if for all $u,v \in \cC$
\begin{equation}
D_\Psi(u, v) \ge \frac{1}{2 \beta} \|u - v\|^2.
\end{equation}
Combining this with \cite[Proposition 3]{fy_losses},
we have for all $\theta \in \Theta$ and all $u \in \cC$
\begin{equation}
S_\Omega(\theta, u) 
\ge D_\Psi(u, \nabla \Omega^*(\theta))
\ge \frac{1}{2\beta} \|u - \nabla \Omega^*(\theta)\|^2.
\end{equation}
Altogether, we thus get for all $\theta \in \Theta$ and $q \in \triangle^{|\cY|}$
\begin{align}
\delta s_\Omega(\theta, q) 
&= S_\Omega(\theta, \muv(q)) \\
&\ge D_\Psi(\muv(q), \nabla \Omega^*(\theta))\\
&\ge \frac{1}{2\beta} \|\muv(q) - \nabla \Omega^*(\theta)\|^2\\
&\ge \frac{1}{8 \beta \sigma^2} \delta \ell(d(\theta), q)^2 \\
&\ge \frac{\epsilon^2}{8 \beta \sigma^2}.
\end{align}
\hfill $\square$

\subsection{Finalizing the proof}

We simply combine \eqref{eq:population_risk_calibration} and Lemma
\ref{lemma:lower_bound}. $\square$

\end{document}